\documentclass{article} 
\usepackage{iclr_arxiv,times}

\usepackage{amsmath,amsfonts,bm}

\def\eqref#1{equation~\ref{#1}}

\def\1{\bm{1}}

\DeclareMathAlphabet{\mathsfit}{\encodingdefault}{\sfdefault}{m}{sl}
\SetMathAlphabet{\mathsfit}{bold}{\encodingdefault}{\sfdefault}{bx}{n}

\usepackage{hyperref}
\usepackage{url}
\usepackage{graphicx}
\usepackage{amsmath, amsfonts, amssymb}
\usepackage{xcolor}
\usepackage{colortbl}
\usepackage{multicol}
\usepackage{multirow}
\usepackage{array}
\usepackage{tabularray}
\usepackage{booktabs}
\usepackage{colortbl}
\usepackage{hhline}
\usepackage{arydshln}
\UseTblrLibrary{booktabs}

\usepackage{adjustbox}
\usepackage{pifont}
\usepackage[misc]{ifsym}
\usepackage{hhline}
\usepackage{makecell}
\usepackage{arydshln}
\usepackage{capt-of}
\usepackage{caption}

\newcommand\Tstrut{\rule{0pt}{2.2ex}}         
\title{GROW: Grounding with Long-Term Awareness for Skill-Incremental Robotic Manipulation}

\author{
\bfseries
Haozhi Cao$^{1}$ \qquad Yasuhiro Kato$^{1}$ \qquad Naoki Morihira$^{1,2}$ \qquad Akinobu Hayashi$^{1,2}$ \\
\bfseries\hspace{40sp} Ziteng Cui$^{1}$ \qquad Yuan Gan$^{1}$ \qquad Tatsuya Harada$^{1,3}$\\
\normalfont
$^{1}$ The University of Tokyo \qquad $^{2}$ Honda R\&D Co., Ltd. \qquad $^{3}$ RIKEN AIP
}

\iclrfinalcopy 
\begin{document}

\maketitle

\begin{abstract}
Skill-incremental robotic manipulation requires continuously acquiring new skills while preserving previously learned capabilities. A widely adopted paradigm formulates manipulation as keyframe-wise action prediction, where the policy predicts the next keyframe action from the current observation and instruction. Applying skill-incremental learning to such keyframe-based manipulation introduces a fundamental level mismatch: local keyframe supervision and replay provide limited explicit guidance for preserving the long-term semantics of complete skill execution. To address this discrepancy, we propose \textbf{GROW} (\textbf{Gr}ounding with L\textbf{o}ng-Term A\textbf{w}areness), a continual manipulation framework that incorporates long-horizon action semantics across perception, action prediction, and replay, enabling skill-level semantics to be better captured and preserved across stages. At the perception level, GROW introduces dual-horizon grounding to regularize cross-modal attention toward both immediate interaction targets and long-horizon task goals. An auxiliary action chain predictor further captures how each skill evolves toward completion through instruction-conditioned future action prediction. The resulting action-chain predictions are additionally exploited for chain-based replay herding, enabling replay selection based on trajectory-level skill semantics. Experiments on RLBench simulation and real-world manipulation tasks show that GROW consistently improves performance over prior skill-incremental learning approaches. Code will be released upon acceptance. 
\end{abstract}

\section{Introduction}
Robotic manipulation~\citep{shridhar2023perceiver,shan2026learning} has attracted increasing attention, benefiting substantially from the rapid development of multi-modal perception~\citep{cao2024reliable,qian2026splatssc}, language-conditioning policy learning~\citep{ze2023gnfactor,zhao2025vlas}, and scalable imitation learning~\citep{kim2024openvla,ni2026swiftvla}. While recent universal policies can handle multiple manipulation tasks~\citep{kim2024openvla,hou2026world}, most assume a predefined set of tasks or skills of interest during training and deployment. Such a paradigm limits the adaptability of intelligent agents, which should be capable of incrementally acquiring new manipulation skills during interaction with human users while retaining the previously learned behaviors. 

To address this gap, incremental learning for robotic manipulation has emerged as a promising solution, with approaches that diverge across their settings and designs. Early attempts, motivated by lifelong-learning benchmarks such as LIBERO~\citep{liu2023libero}, mainly formulate continual manipulation as sequential adaptation to individual tasks with variations in objects, spatial configurations, task goals, or their compositions. Although such variations may substantially change observations or instructions, many successive tasks share common interaction patterns despite variations. Various learning-without-forgetting mechanisms have been explored, such as segment discovery~\citep{wan2024lotus}, task-specific parameter tuning~\citep{liu2024tail}, and latent replay~\citep{roy2025m2distill,yu2026lifelong}. More recently, iManip~\citep{zheng2025imanip} considers a more behaviorally diverse incremental setting named skill-incremental learning (SIL), where successive stages introduce distinct manipulation skills (e.g., ``close the jar'' vs. ``open the drawer'') that may require different interaction patterns and action primitives. To accommodate such skill expansion, iManip leverages keyframe-based manipulation policies~\citep{shridhar2023perceiver,lu2024manigaussian} that sparsify dense trajectories into critical keyframe states and formulate manipulation as sequential next-keyframe prediction.

\begin{figure}
    \centering
    \includegraphics[width=\linewidth]{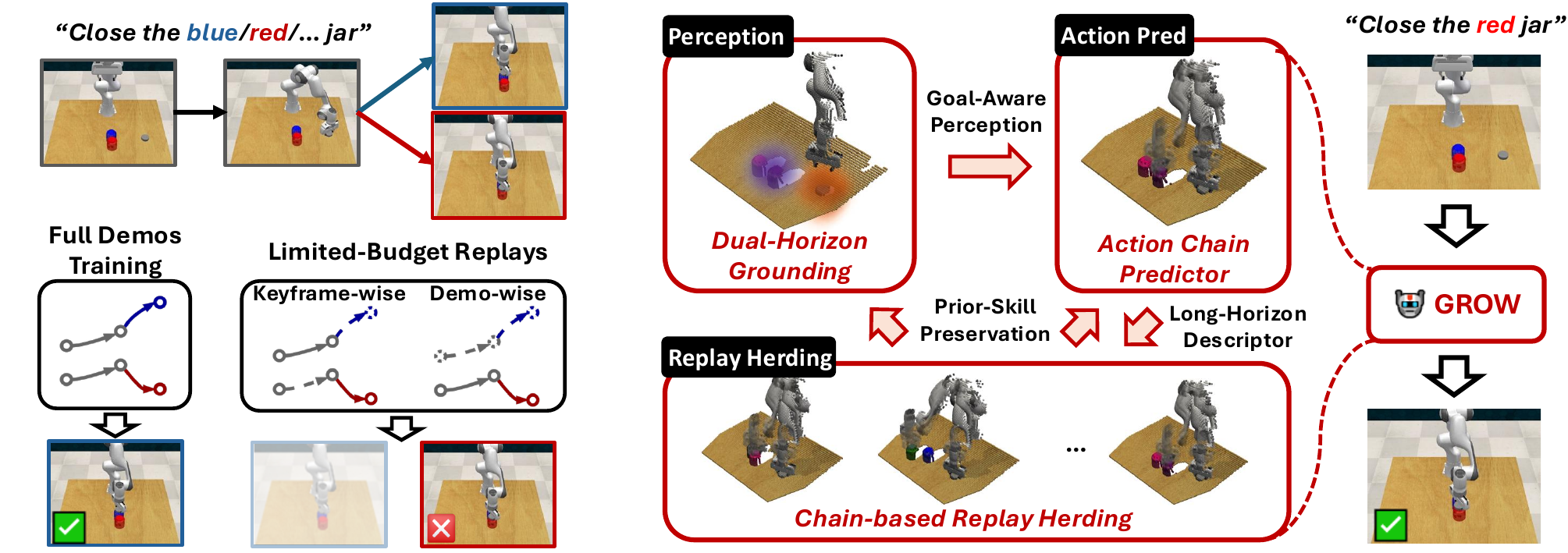}
    \caption{Illustration of the level mismatch in keyframe-based SIL (left) and GROW (right). Taking ``close the \textcolor{blue}{blue} /\textcolor{red}{red} jar'' as an example, the original next-keyframe prediction may provide limited guidance for preserving its instruction-conditioned execution contexts. GROW is designed to preserve such long-horizon semantics throughout perception, action prediction, and replay herding.}
    \label{fig:intro}
    \vspace{-10pt}
\end{figure}

While such a compact formulation is well suited for representing diverse manipulation behaviors, directly extending it to SIL exposes a critical level mismatch: \emph{learning and replay operate on individual keyframes, whereas the knowledge to preserve spans complete skill execution.} Successful manipulation requires preserving how intermediate interactions relate to the instruction-conditioned task goal, whereas keyframe-based SIL optimizes and rehearses these interactions primarily through local next-keyframe prediction. For instance, ``close the \textcolor{blue}{blue}/\textcolor{red}{red} jar'' may share similar local motions when picking up the lid, while correctly associating these actions with the instructed target requires preserving their long-horizon execution context (see Figure~\ref{fig:intro}). This limitation is less pronounced when complete demonstrations of all skills are jointly available, as the policy is repeatedly exposed to keyframes spanning different stages of each instruction-conditioned execution. Under SIL, however, previous demonstrations are reduced to a limited replay subset. Under a fixed replay budget, keyframe-wise replay~\citep{zheng2025imanip} provides broader coverage of local states, actions, and instruction variants, but retains only sparse fragments of each skill execution. Conversely, replaying complete demonstrations preserves execution continuity at the cost of substantially reduced sample diversity. This trade-off motivates preserving long-horizon execution semantics without requiring full-trajectory replay.

Motivated by this discrepancy, we argue that SIL should preserve not only local action mappings, but also the long-horizon semantics relating current interactions to subsequent skill progression and the eventual task goal. To this end, we introduce \textbf{Gr}ounding with L\textbf{o}ng-Term A\textbf{w}areness (\textbf{GROW}), which augments conventional keyframe-based manipulation with long-horizon semantics to better capture and preserve skill-level execution knowledge across incremental stages. Specifically, GROW incorporates long-term awareness across perception, action prediction, and replay herding. At the perception level, dual-horizon grounding is introduced to regularize the early cross-attention layer towards both immediate interaction regions and long-term task goals. Beyond these isolated spatial anchors, an action chain predictor models the instruction-conditioned future progression of each skill, providing long-horizon execution context for immediate action prediction. The resulting action-chain representations are further exploited for chain-based replay herding, enabling representative replay samples to be selected based on trajectory-level execution patterns instead of local actions. To summarize, our contributions are as follows:
\begin{itemize}
    \item We identify a critical mismatch between keyframe-level learning and skill-level knowledge preservation in skill-incremental manipulation, motivating long-horizon semantic modeling beyond local action mappings.
    
    \item We propose GROW, which introduces long-term awareness into perception, action prediction, and replay herding. Together, it enables long-horizon execution knowledge to be explicitly captured and propagated across incremental stages.
    
    \item Extensive experiments on RLBench across three manipulation backbones show that GROW consistently outperforms existing incremental-learning baselines. Real-world experiments further demonstrate its applicability beyond simulation.

\end{itemize}

\section{Related Works}

\textbf{Incremental Learning for Robotic Manipulation.}
The existing incremental learning approaches for robotic manipulation can be divided into two types based on their problem formulation. Motivated by the preliminary exploration of life-long learning proposed in LIBERO~\citep{liu2023libero}, the major works define the continual manipulation problem as a sequential adaptation of individual tasks with controllable variations, such as objects, spatial locations, goals, or their combinations. Based on these settings, multiple attempts have been explored mainly based on existing incremental strategies~\citep{gao2021cril,liu2023libero,wan2024lotus,zhu2022bottom,liu2024tail,lee2024incremental,roy2025m2distill,yu2026lifelong}. Another line considers skill-incremental learning (SIL), which is primarily introduced by iManip~\citep{zheng2025imanip}. For SIL, distinct skills are acquired sequentially, and iManip mitigates skill forgetting through temporally balanced keyframe replay and extendable action prompts. Different from these approaches, GROW focuses on the level mismatch inherent to keyframe-based skill-incremental learning, explicitly preserving long-horizon execution semantics across perception, action prediction, and replay.

\textbf{Perception Grounding for Robotic Manipulation.}
Task-relevant visual grounding has been widely explored to improve manipulation accuracy, which is usually leveraged with Vision-Language-Action (VLA) models. Some approaches leverage specialized external models, such as RoboGround~\citep{huang2025roboground} and VIP~\citep{li2025vip} leveraging LISA~\citep{lai2024lisa} and YOLOv11~\citep{khanam2024yolov11}, respectively. Another branch is to take advantage of the powerful visual-language reasoning ability of VLA, Vision-Language Models (VLMs), or Large Language Models (LLMs)~\citep{zawalski2024robotic,song2026reconvla}. In contrast, GROW utilizes an immediate interaction target and an eventual execution goal inherent in the training data to regularize the internal cross-modal attention without involving external supervision signals or large models. This aims to rectify early perception for both local action prediction and long-horizon skill execution, which is particularly important for knowledge preservation under incremental learning.

\textbf{Long-Horizon Execution Modeling.}
Modeling multiple future actions has also proven effective for robotic manipulation. Representative works include ACT~\citep{zhao2023learning}, which adopts action chunking to achieve affordable smooth actions in fine-grained control, and diffusion-based policies~\citep{chi2023diffusion,xian2023chaineddiffuser,ke20243d}. Other approaches directly synthesize sequences of end-effector waypoints or decompose long-horizon tasks into intermediate interaction steps~\citep{mees2022matters,ni2024generate,jain2025enabling}. These approaches primarily employ hierarchical execution structures for action generation or control. In contrast, GROW retains the next-keyframe prediction policy, leveraging action-chain prediction as an auxiliary supervision signal to encode instruction-conditioned future progression for better skill-level knowledge preservation. 

\section{Methodology}
\label{Sec:method}
\begin{figure}[t]
    \centering
    \includegraphics[width=\linewidth]{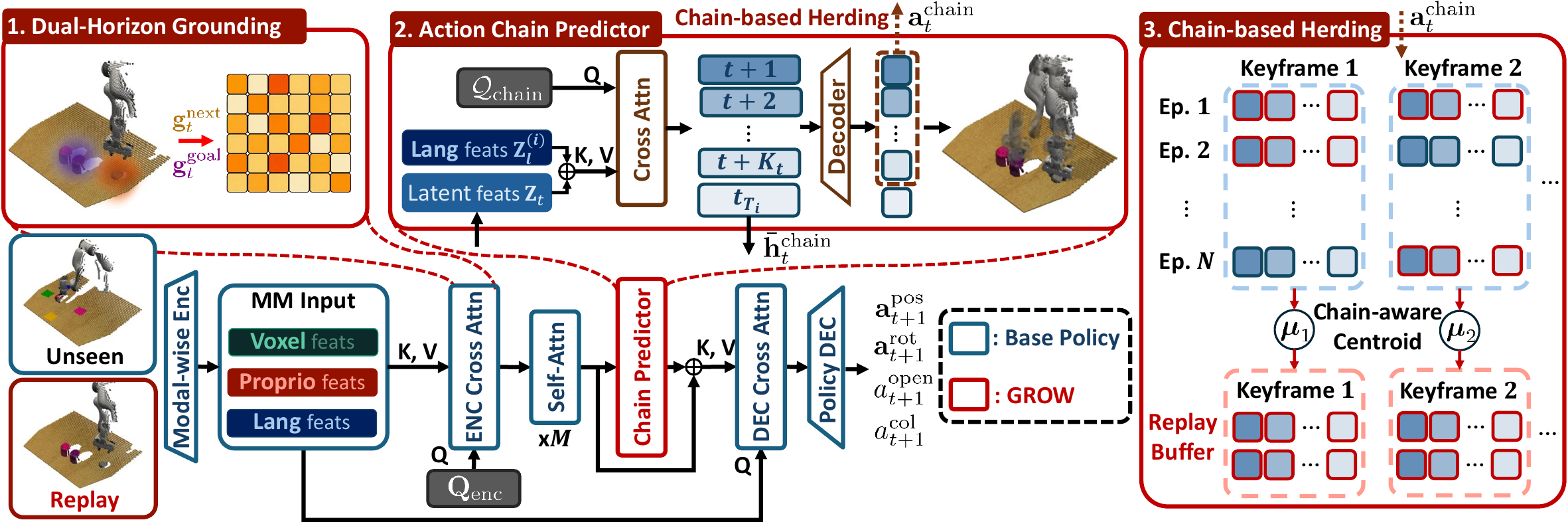}
    \caption{Overall structure of GROW. }
    \label{Fig:main_method}
    \vspace{-10pt}
\end{figure}

In this section, we formulate the problem of SIL and then introduce the architecture of our proposed GROW, a PerceiverIO-based~\citep{jaegle2021perceiver} framework for skill-incremental manipulation. GROW aims to preserve the skill-level knowledge by incorporating long-term action awareness across perception, action prediction, and replay, as illustrated in Figure~\ref{Fig:main_method}. At the perception level, dual-horizon grounding regularizes early cross-modal attention toward task-relevant regions. At the action level, an action chain predictor captures long-horizon execution semantics through multi-step action prediction, whose representations are further exploited for Chain-based Replay Herding. For incremental updates, we adopt stage-wise LoRA with initialization, merging, and calibration, maintaining a fixed model capacity.

\textbf{Problem Formulation.}
Following existing protocol~\citep{ze2023gnfactor,lu2024manigaussian}, we formulate the manipulation task as a keyframe-prediction problem. Specifically, a dataset $\mathcal{D}=\{\zeta_1, \zeta_2,..., \zeta_n\}$ contains $n$ expert demonstrations, each of which is composed of keyframes when joint velocities approach zero while the gripper's openness remains consistent. Each demonstration $\zeta_i=\{\mathcal{A}_i,\mathcal{O}_i, \mathcal{P}_i, \mathbf{l}_i\}$ contains a sequence of actions $\mathcal{A}_i=\{\mathbf{a}_t\}_{t=1}^{T_i}$, observations $\mathcal{O}_i=\{\mathbf{o}_t\}_{t=1}^{T_i}$, proprioception states $\mathcal{P}_i=\{\mathbf{p}_t\}_{t=1}^{T_i}$, and the linguistic instruction $\mathbf{l}_i$ of the $i$-th demonstration, respectively. $\mathbf{a}_t=\{\mathbf{a}^\mathrm{pos}_t, \mathbf{a}^\mathrm{rot}_t, a^\mathrm{open}_t, a^\mathrm{col}_t\}$ denotes an action associated with keyframe $t$, including the gripper position $\mathbf{a}^\mathrm{pos}_t$ and rotation $\mathbf{a}^\mathrm{rot}_t$, gripper openness $a^\mathrm{open}_t\in[0,1]$, and collision avoidance $a^\mathrm{col}_t\in[0,1]$, while an observation $\mathbf{o}_t\in\mathbb{R}^{H\times W\times L\times3}$ is captured as the colorful 3D voxels from an RGB-D image. $\mathbf{p}_t\in\mathbb{R}^4$ denotes the gripper state composed of the finger positions and openness. The model $\phi(\cdot)$ is trained to predict actions $\hat{\mathbf{a}}_{t+1}=\phi(\mathbf{o}_t,\mathbf{p}_t, \mathbf{l}_i)$.

Under the skill-incremental setting~\citep{zheng2025imanip}, learning proceeds over a sequence of stages $s=1,\dots, S$, where each stage introduces an unseen dataset $\mathcal{D}^{(s)}_\mathrm{new}$ corresponding to some novel manipulation skills with interaction semantics distinct from the previously learned skills (e.g., stacking blocks vs. closing a jar). Starting from the model learned at $s-1$, the objective is to obtain a model $\phi^{(s)}$ that acquires the new skills while retaining competence on all the previously learned skills.

\subsection{Keyframe-based PerceiverIO Backbones}
Existing keyframe-based manipulation approaches, such as PerACT~\citep{shridhar2023perceiver}, GNFactor~\citep{ze2023gnfactor}, and ManiGaussian~\citep{lu2024manigaussian}, commonly adopt PerceiverIO-style architectures to aggregate multi-modal observations and predict subsequent actions. At keyframe $t$, observation, linguistic, and proprioceptive features are first flattened,  encoded, and concatenated into a multi-modal representation $\mathbf X_t$. The Perceiver encoder employs a set of latent queries $\mathbf Q_{\mathrm{enc}}$ to aggregate multi-modal information through cross attention~\citep{vaswani2017attention}, denoted as:
\begin{equation}
    \mathbf{Z}^0_t = \texttt{CrossAttn}_\mathrm{enc}(\mathbf Q_{\mathrm{enc}}, \mathbf{X}_t, \mathbf{X}_t),\label{Eq:enc_crossattn}
\end{equation}
where $\mathbf{Q}_\mathrm{enc}$ is the query input, while $\mathbf{X}_t$ serves as the keys and values input; all are encoded with input-specific linear weights before the attention computation. The extracted latent features $\mathbf{Z}^0_t$ are then aggregated by several self-attention blocks, whose output is queried by the multi-modal representation $\mathbf{X}_t$, formulated as:
\begin{equation}
    \mathbf{Z}_t=\texttt{SelfAttn}_{\times M}(\mathbf{Z}^0_t),\quad \mathbf{H}_t=\texttt{CrossAttn}_\textrm{dec}(\mathbf{X}_t, \mathbf{Z}_t, \mathbf{Z}_t),\label{Eq:sa_features}
\end{equation}
where $\mathbf{Z}_t$ denotes the last latent representation before the cross-attention decoder. $\mathbf{H}_t$ is subsequently leveraged to decode action $\hat{\mathbf{a}}_{t+1}$ through the action decoder.

\textbf{Discussion.} Despite the architectural difference, existing PerceiverIO-based methods optimize each observation primarily for next-keyframe prediction. As in Figure~\ref{fig:intro}, such local supervision and keyframe-level replay~\citep{zheng2025imanip} provide limited explicit guidance on how long-horizon dependencies connect individual interactions into coherent skill execution, motivating GROW.

\subsection{GROW: Grounding with Long-Term Awareness}
\textcolor{black}{To address this limitation and enable long-horizon semantic propagation across incremental stages, we introduce GROW, which augments conventional keyframe-wise learning with long-term action awareness, allowing skill-level semantics (i.e., which targets are involved and in what order) to be explicitly maintained across perception, action prediction, and replay.}

\textbf{Dual-Horizon Grounding.}
Perception is an important preliminary to provide informative clues for long-horizon skill execution. However, we observe (Figure~\ref{fig:grounding_viz}) that the encoder cross-attention of existing PerceiverIO-based policies can remain diffuse over task-irrelevant regions, and such grounding becomes increasingly fragile across incremental stages. To reinforce the connection between perception and skill execution, GROW regularizes the encoder cross-attention through two complementary spatial anchors, including the next-keyframe position for immediate interaction and the terminal-keyframe position for long-horizon task context. Specifically, given the input multi-modal representation $\mathbf{X}_t$, the attention map of $\texttt{CrossAttn}_\mathrm{enc}$ in Equation~\ref{Eq:enc_crossattn} is preserved as:
\begin{equation}
    \mathbf{A}_t = \mathrm{Softmax}(\mathbf{Q}_{\mathrm{enc}}\mathbf{W}_Q(\mathbf{X}_t\mathbf{W}_K)^\top / \sqrt{d}),
\end{equation}
where $\mathbf{W}_Q$ and $\mathbf{W}_K$ denote the learnable query and key projections of the encoder cross-attention. Given $\mathbf{A}_t$ as the attention map between $\mathbf{Q}_\mathrm{enc}$ and $\mathbf{X}_t$, since $\mathbf{X}_t$ contains both spatial observations and non-spatial modalities, we retain only the attention scores correlated to the voxelized observation $\mathbf{o}_t$ through mask selection, aggregate them across latent queries and attention heads, and normalize the results to form an attention distribution $\alpha_t$ as:
\begin{equation}
    \alpha_t = \mathrm{Norm}(\mathrm{Agg}(\mathbf{A}_t[:,\mathcal{I}_\mathrm{voxel}])),
\end{equation}
where $\mathcal{I}_\mathrm{voxel}$ denotes the indices of voxel tokens in $\mathbf{X}_t$ and $\mathrm{Norm(\cdot)}$ indicates the normalization across all selected elements. The next-step and goal grounding anchors are then formulated as 3D Gaussian distributions centered at the spatial location related to the immediate action $\mathbf{a}_{t+1}$ and goal action $\mathbf{a}_{T_i}$, formulated as:
\begin{equation}\label{Eq:ground_target}
    \mathbf{g}^{\textrm{next}}_t = \mathrm{Norm}(\mathcal{N}(\mathbf{a}^\mathrm{pos}_{t+1}, \sigma^2\mathbf{I})), \quad
    \mathbf{g}^{\textrm{goal}}_t = \mathrm{Norm}(\mathcal{N}(\mathbf{a}^\mathrm{pos}_{T_i}, \sigma^2\mathbf{I})),
\end{equation}
where $\sigma$ denotes the predefined variance of Gaussian distributions for both next-step and goal grounding anchors. The dual-horizon grounding objective is then formulated as:
\begin{equation}\label{Eq:grounding_loss}
    \mathcal{L}_\mathrm{dh}=\lambda_\mathrm{n} D_\mathrm{KL}(\mathbf{g}^{\textrm{next}}_t \| \alpha_t) +
                            \lambda_\mathrm{g} D_\mathrm{KL}(\mathbf{g}^{\textrm{goal}}_t \| \alpha_t),
\end{equation}
where $\lambda_\mathrm{n}$ and $\lambda_\mathrm{g}$ are the loss coefficients for immediate and long-horizon grounding, respectively. The former, next-keyframe grounding, encourages the model to focus on the immediate interaction region based on which to predict the next-step action, while the latter goal grounding supplies the long-horizon goal of where the current action ultimately leads, complementing long-horizon context in the early perception stage. 

\textbf{Action Chain Predictor.}
Dual-horizon grounding preserves semantic cues about both the immediate interaction and terminal goal, but they remain isolated spatial targets, lacking explicit modeling of how execution evolves between them. We therefore introduce an auxiliary action chain predictor that encodes future progression of each skill. 

Such future progression is highly instruction-conditioned (Figure~\ref{fig:intro}). Although the instruction $\mathbf{l}_i$ has been incorporated during the early stage of the Perceiver representation, we explicitly reintroduce its representation to provide direct instruction-specific conditioning for long-horizon action chain prediction. Given the latent features extracted by the last self-attention layers $\mathbf{Z}_t$ as in Eq.~\ref{Eq:sa_features}, the action chain predictor first concatenates them with the re-projected linguistic representation as:
\begin{equation}
    \mathbf{Z}_\mathrm{l}^{(i)}=\mathbf{E}^{(i)}_\mathrm{l}\mathbf{W}_\mathrm{l}, \quad
    \bar{\mathbf Z}_t = \mathrm{MeanPool}(\mathbf Z_t), \quad
    \tilde{\mathbf{Z}}_t = \mathbf{Z}_\mathrm{l}^{(i)} \oplus \bar{\mathbf Z}_t,
\end{equation}
where $\mathbf{E}^{(i)}_\mathrm{l}$ denotes the previously encoded linguistic representation existing in $\mathbf{X}_t$ and $\mathrm{MeanPool}$ indicates the average pooling along the token dimension. $\mathbf{W}_\mathrm{l}$ represents the re-projection in the action chain predictor. The resulting $\tilde{\mathbf{Z}}_t$ subsequently serves as the keys and values for a set of $k$ learnable chain queries $\mathcal{Q}_\mathrm{chain}=\{\mathbf{q}_j\}_{j=1}^{k}$ to extract the action chain representation $\mathbf{H}^\mathrm{chain}_t$ as:
\begin{equation}\label{Eq:chain_xA}
    \mathbf{H}^\mathrm{chain}_t = \texttt{SelfAttn}_\mathrm{chain}(\texttt{CrossAttn}_\mathrm{chain}(\mathcal{Q}_\mathrm{chain}, \tilde{\mathbf{Z}}_t,\tilde{\mathbf{Z}}_t)).
\end{equation}
The valid prediction steps depend on the remaining steps in the trajectory. Specifically, given $R_t = T_i-t$ denoting the number of remaining keyframes, we define $K_t = \min(R_t-1, k-1)$ to retrieve the first $K_t+1$ representations in $\mathbf{H}^\mathrm{chain}_t$ and predict the corresponding positions using an MLP-based action decoder. The first $K_t$ representations predict positions of consecutive future keyframes, while the $(K_t+1)$-th representation predicts the position of the goal, denoted as: 
\begin{equation}
    \mathcal{A}_t^\mathrm{chain} = \{\hat{\mathbf{a}}_{t+1}^\mathrm{pos},\hat{\mathbf{a}}_{t+2}^\mathrm{pos},...\hat{\mathbf{a}}_{t+K_t}^\mathrm{pos}, \hat{\mathbf{a}}_{T_i}^\mathrm{pos}\}.
\end{equation}
Given the action chain representation and predictions, an exponentially decaying weight $w^\mathrm{pred}_j=\tau^{j-1}/\sum^{K_t+1}_{r=1}\tau^{r-1}$ controlled by the decay rate $\tau\in(0,1]$ prioritizes the near-future execution while retaining long-horizon task information. The chain predictor is trained with the objective as:
\begin{equation}\label{Eq:chain_loss}
\mathcal{L}_\mathrm{chain}
=
\sum_{j=1}^{K_t}
w^\mathrm{pred}_j
\mathcal{L}_\mathrm{pos}
\left(
\hat{\mathbf{a}}_{t+j}^{\mathrm{pos}},
\mathbf{a}_{t+j}^{\mathrm{pos}}
\right)+w^\mathrm{pred}_{K_t+1}
\mathcal{L}_\mathrm{pos}
\left(
\hat{\mathbf{a}}_{T_i}^{\mathrm{pos}},
\mathbf{a}_{T_i}^{\mathrm{pos}}
\right),
\end{equation}
where $\mathcal{L}_\mathrm{pos}$ denotes the identical pose loss function used for the next-keyframe prediction. $\mathbf{H}^\mathrm{chain}_t$ is then detached and injected into the backbone to provide long-term awareness as:
\begin{equation}\label{Eq:chain_feat}
\bar{\mathbf{h}}_t^{\mathrm{chain}} = \sum_{j=1}^{k} w^\mathrm{feat}_j\mathbf{h}_{t,j}^{\mathrm{chain}},\quad 
\mathbf{Z}_t^{+} = \texttt{concat}(\mathbf{Z}_t, \bar{\mathbf{h}}_t^{\mathrm{chain}}),
\end{equation}
where $w^\mathrm{feat}_j=\tau^{j-1}/\sum^{k}_{r=1}\tau^{r-1}$ is the feature decaying weight controlled by the same decay factor $\tau$. $\mathbf{h}_{t,j}^{\mathrm{chain}}$ denotes the corresponding $j$-th chain representation in $\mathbf{H}^\mathrm{chain}_t$, while $\mathbf{Z}_t^{+}$ is the long-term aware representation that replaces $\mathbf{Z}_t$ during action decoding in Eq.~\ref{Eq:sa_features}. While only the first $K_t+1$ query outputs receive direct supervision for a given keyframe, all $k$ queries are shared across training samples and are supervised whenever their corresponding slots become valid. We therefore aggregate the complete query representation as a fixed-dimensional long-horizon context for immediate action prediction.

\textbf{Chain-based Replay Herding.}
Experience replay is widely adopted in incremental learning to mitigate catastrophic forgetting, where a limited subset of samples from previous stages is retained and replayed during subsequent training. Under keyframe-based manipulation, existing approaches~\citep{zheng2025imanip} propose to utilize per-keyframe prediction statistics or features as the selection criteria. Despite being consistent with the next-keyframe prediction paradigm, such strategies categorize each keyframe sample largely based on its immediate movement, providing limited information about how the remaining manipulation evolves for task completeness. In contrast, GROW implements a chain-based replay herding that leverages the long-horizon action information from the chain predictor to achieve replay herding with long-term awareness. Taking a keyframe sample $t$ as an example, given the predicted chain logits $\mathcal{A}_t^\mathrm{chain}$, we compute its corresponding long-term prediction descriptors through pooling using the same $w^\mathrm{pred}_j$, formulated as:
\textcolor{black}{
\begin{equation}\label{Eq:lt_descriptoer}
    \mathbf{a}^\mathrm{chain}_t = \sum_{j=1}^{K_t}
w^\mathrm{pred}_j
\hat{\mathbf{a}}_{t+j}^{\mathrm{pos}}+w^\mathrm{pred}_{K_t+1}
\hat{\mathbf{a}}_{T_i}^{\mathrm{pos}}, \quad
\hat{\mathbf{a}}^\mathrm{chain}_t = \frac{\mathbf{a}^\mathrm{chain}_t}{\lVert \mathbf{a}^\mathrm{chain}_t \rVert_2},
\end{equation}}
For replay herding, we conduct sample selection independently within each
previously learned task. Let $\tilde{\mathcal{D}}=\{\zeta_i\}_{i=1}^{N}$ denote the demonstration subset belonging to the current task, and $M$ denote the representative trajectory length computed as the average keyframe number of $\zeta_i$. For each keyframe position $m\in\{1,\ldots, M\}$, we compute the corresponding chain-aware centroid by averaging the long-term prediction descriptors of these keyframes:
\textcolor{black}{
\begin{equation}
    \boldsymbol{\mu}_m =
    \frac{1}{|\mathcal{I}_m|}
    \sum_{i\in\mathcal{I}_m}
    \hat{\mathbf{a}}^{\mathrm{chain}}_{i,m},
\end{equation}}
where $\mathcal{I}_m$ contains the indices whose corresponding demonstrations contains at least $m$ keyframes. \textcolor{black}{$\hat{\mathbf{a}}^{\mathrm{chain}}_{i,m}$} denotes the pooled action-chain logit of the $m$-th keyframe from demonstration $\zeta_i$ as in Eq.~\ref{Eq:lt_descriptoer}. Following mean-based herding~\citep{rebuffi2017icarl,castro2018end}, candidate samples are ranked according to their Euclidean distance to the corresponding centroid \textcolor{black}{$d_{i,m} =\left\|\hat{\mathbf{a}}^{\mathrm{chain}}_{i,m} - \boldsymbol{\mu}_m\right\|_2$}.
Given a budget of $B_m$ samples at position $m$, we retain the candidates with the smallest distances:
\begin{equation}\label{Eq:chain_replay}
    \mathcal{R}_m =
    \operatorname*{arg\,min}_{
        \mathcal{S}\subseteq\mathcal{I}_m
        }
    \sum_{i\in\mathcal{S}} d_{i,m}, \quad \text{s.t. } |\mathcal{S}| = B_m .
\end{equation}
The whole replay indices for the task are therefore represented as $\mathcal{R} =\bigcup_{m=1}^{M}\left\{(m, \mathcal{R}_m)\right\}$. For each replay index, we store the corresponding keyframe input, its next-step ground-truth action, and the valid future ground-truth action positions required to supervise the action chain predictor. These replay samples are then stochastically mixed with new-task data during incremental training.

\subsection{Incremental Update and Objectives.}
For each incremental stage, we employ a LoRA-based~\citep{hu2021lora} update strategy to update the shared backbone with a constrained number of trainable parameters. To maintain a constant adaptation budget while preserving sufficient plasticity for newly introduced skills, GROW further adopts stage-wise LoRA initialization, merging, and calibration. Specifically, a fresh set of LoRA parameters is initialized at the beginning of each stage and optimized along with the chain predictor and the PerceiverIO's action decoder, while the other components in PerceiverIO remain frozen. After the training process, the LoRA components are merged back into the backbone through summation ($\mathbf{W}=\mathbf{W}_0 + \mathbf{B}\mathbf{A}\alpha/r$ as in~\citep{hu2021lora}), followed by a short calibration fine-tuning using the same objectives and training data but a reduced learning rate. Details about LoRA-targeted components and implementations are included in our appendix.

The overall learning objectives consist of four components. First, a conventional action loss is adopted to facilitate the imitation learning of the backbone, formulated as:
\begin{equation}
    \mathcal{L}_\mathrm{action}=
    \mathcal{L}_\mathrm{pos}(\hat{\mathbf{a}}^\mathrm{pos}_{t+1}, \mathbf{a}^\mathrm{pos}_{t+1}) + 
    \mathcal{L}_\mathrm{rot}(\hat{\mathbf{a}}^\mathrm{rot}_{t+1}, \mathbf{a}^\mathrm{rot}_{t+1}) +
    \mathcal{L}_\mathrm{open}(\hat{a}^\mathrm{open}_{t+1}, a^\mathrm{open}_{t+1}) +
    \mathcal{L}_\mathrm{col}(\hat{a}^\mathrm{col}_{t+1}, a^\mathrm{col}_{t+1}),
\end{equation}
where $\hat{\mathbf{a}}^\mathrm{pos}_{t+1}, \hat{\mathbf{a}}^\mathrm{rot}_{t+1}, \hat{a}^\mathrm{open}_{t+1}, \hat{a}^\mathrm{col}_{t+1}$ denote the backbone's next-keyframe predictions of positions, rotations, collision, and openness, respectively. In practice, the loss function for each component is implemented as cross-entropy loss following the protocol of previous works~\citep{lu2024manigaussian,zheng2025imanip}. Additionally, a knowledge distillation loss is applied to the replay samples to better preserve previously acquired knowledge using the model from the previous stage as the teacher:
\begin{equation}
    \mathcal{L}_\mathrm{dis}=
    \mathrm{MSE}(\hat{\mathbf{a}}^\mathrm{pos}_{t+1}, \tilde{\mathbf{a}}^\mathrm{pos}_{t+1}) +
    \mathrm{MSE}(\hat{\mathbf{a}}^\mathrm{rot}_{t+1}, \tilde{\mathbf{a}}^\mathrm{rot}_{t+1}) +
    \mathrm{MSE}(\hat{a}^\mathrm{open}_{t+1}, \tilde{a}^\mathrm{open}_{t+1}) + 
    \mathrm{MSE}(\hat{a}^\mathrm{col}_{t+1}, \tilde{a}^\mathrm{col}_{t+1}),
\end{equation}
where $\tilde{\mathbf{a}}^\mathrm{pos}_{t+1}, \tilde{\mathbf{a}}^\mathrm{rot}_{t+1}, \tilde{a}^\mathrm{open}_{t+1}, \tilde{a}^\mathrm{col}_{t+1}$ denote the corresponding predictions from the trained model of the previous stage. With the GROW-specific objectives, the total loss function can be represented as:
\begin{equation}
    \mathcal{L}_\mathrm{total} = 
    \mathcal{L}_\mathrm{action} + 
    \lambda_\mathrm{dis}\mathcal{L}_\mathrm{dis} + 
    \lambda_\mathrm{chain}\mathcal{L}_\mathrm{chain} +
    \mathcal{L}_\mathrm{dh},
\end{equation}
where $\lambda_\mathrm{dis},\lambda_\mathrm{chain}$ are the coefficients for the distillation loss and chain prediction loss, respectively.

\section{Experimental Results}\label{sec:exp}

\textbf{Benchmark Details.}
Following existing works~\citep {lu2024manigaussian,zheng2025imanip}, our experiments are conducted on 10 manipulation skills in RLBench~\citep{james2020rlbench}, supplemented with real-world rollouts involving 3 skills. For simulation, each skill contains 20 demonstrations sampled from at least 2 variations, while the evaluation involves 25 episodes per task. Following the SIL protocol of iManip, we utilize the same configuration notation ``B\textit{n}-\textit{k}N\textit{m}'', which indicates agents are pre-trained on \textit{n} base skills and incrementally updated across \textit{k} stages, each of which introduces \textit{m} unseen skills. For real-world rollouts, we collect 20 demonstration episodes as training data and evaluate each method with 10 rollouts. More details are included in our appendix.

\textbf{Baseline Details.}
GROW is mainly compared with \textbf{iManip}~\citep{zheng2025imanip}, the existing state-of-the-art (SOTA) keyframe-based SIL method, and the baseline used in their comparison, called \textbf{TIB}~\citep{rebuffi2017icarl}. We also adopt 3 baselines mostly adopted in lifelong learning~\citep{liu2023libero}, including sequential fine-tuning (\textbf{Seq-FT}), fine-tuning with a Fisher-information-based regularizer (\textbf{EWC})~\citep{kirkpatrick2017overcoming}, and \textbf{TAIL}~\citep{liu2024tail,yu2026lifelong}. We include two variants of TAIL. \textcolor{black}{TAIL (oracle) is the original formulation~\citep{liu2024tail}, which trains and activates a task-specific LoRA adapter for each skill using the ground-truth task identity. Since each skill is evaluated with its own adapter (or the base policy for base skills), it is free from forgetting by construction, which serves as an oracle reference and is excluded from the comparison. TAIL-A is the task-agnostic variant introduced by \citet{yu2026lifelong}, following LOTUS~\citep{wan2024lotus}, which requires no task identity: the success rate of each skill is averaged over all adapters available at the current stage, simulating inference without task identity.} Our main comparison is conducted across 3 keyframe-based backbones, including PerAct~\citep{shridhar2023perceiver}, GNFactor~\citep{ze2023gnfactor}, and ManiGaussian~\citep{lu2024manigaussian}. The other incremental settings and ablation studies in simulation are based on ManiGaussian, while the real-world rollouts are all based on PerAct. \emph{Note that we utilize our self-implemented version of iManip and TIB, as their code is not available till submission}. More details of baselines are in our appendix.

\textbf{Implementation Details and Metrics.}
All experiments are conducted on an NVIDIA RTX PRO 6000 GPU using 100k iterations for base pre-training and 50k iterations for each incremental stage. Most input preprocessing protocols follow existing works~\citep{shridhar2023perceiver,ze2023gnfactor,lu2024manigaussian}. For each incremental stage, we compute the success rate of tasks from previous stages (Old), tasks newly introduced (New), and overall average (All). More details are in our appendix.

\begin{table}[t]
\centering
\huge
\setlength{\tabcolsep}{4pt}
\caption{\textcolor{black}{Stage-wise performance comparison under B5-5N1. The first column indicates the backbone (BB) used for each row, including PerAct (PA), GNFactor (GF), and ManiGaussian (MG). \textcolor{gray}{TAIL} is an oracle reference using test-time task identity, excluded from the comparison and bolding.}}
\vspace{-3pt}
\resizebox{.9\textwidth}{!}{
\begin{tblr}{ 
        colspec             = {c | l | c *{25}{c}},
        cell{1}{1,2} = {r=2}{l},
        cell{1}{3, 4, 8, 12, 16, 20, 24}        = {r=2}{c}, 
        cell{1}{5, 9, 13, 17, 21, 25} = {c=3}{c},
        cell{3, 10, 17}{1}         = {r=7}{c},
        cell{9, 16, 23}{2-Z} = {bg=gray!25},
        cell{3, 10, 17}{2-Z} = {fg=gray}
    }
\hline
BB & Method & Base & & Stage 1 & & & & Stage 2 & & & & Stage 3 & & & & Stage 4 & & & & Stage 5 & & & & Avg \\
\hline
& & & & Old & New & All & & Old & New & All & & Old & New & All 
   & & Old & New & All & & Old & New & All & & Old & New & All\Tstrut\\
\hline

\rotatebox[origin=c]{90}{PA} 
& TAIL
& \textbf{34.4} & & 40.8 & 56.0 & 43.3 & &
        43.3 & 20.0 & 40.0 & &
        40.0 & 16.0 & 37.0 & &
        38.5 & 12.0 & 35.6 & &
        33.3 & 12.0 & 31.2 & &
        39.2 & 23.2 & 37.4\Tstrut\\

\hline

& Seq-FT 
& \textbf{34.4} & & 10.4 & 56.0 & 18.0 & & 
           5.3 & 8.0 & 5.7 & & 
           4.6 & 16.0 & 6.0 & & 
           1.0 & \textbf{28.0} & 4.0 & & 
           3.6 & \textbf{24.0} & 5.6 & & 
           5.0 & \textbf{26.4} & 7.9\Tstrut\\
& EWC 
& \textbf{34.4} & & 14.4 & \textbf{68.0} & 23.3 & & 
           6.0 & 0.0 & 5.1 & & 
           4.0 & 4.0 & 4.0 & & 
           0.5 & 8.0 & 1.3 & & 
           2.7 & 0.0 & 2.4 & & 
           5.5 & 16.0 & 7.2 \\
& TAIL-A 
& \textbf{34.4} & & 25.6 & 12.0 & 23.3 & & 
           18.7 & 1.3 & 16.2 & & 
           12.9 & 3.0 & 11.6 & & 
           11.6 & 4.8 & 10.8 & & 
           8.8 & 0.0 & 7.9 & & 
           15.5 & 4.2 & 14.0\\
\hline[dashed]
& TIB
& \textbf{34.4} & & 28.0 & 36.0 & 29.3 & & 
           16.0 & 0.0 & 13.7 & & 
           \textbf{24.6} & 16.0 & \textbf{23.5} & & 
           14.0 & 8.0 & 13.3 & & 
           15.6 & 0.0 & 14.0 & & 
           19.6 & 12.0 & 18.8 \Tstrut\\
& iManip 
& \textbf{34.4} & & 24.0 & 16.0 & 22.7 & & 
           22.7 & 12.0 & 21.1 & &
           17.1 & 4.0 & 15.5 & &
           \textbf{19.5} & 12.0 & \textbf{18.7} & &
           \textbf{18.7} & 0.0 & \textbf{16.8} & &
           20.4 & 8.8 & 19.0 \\
& GROW 
& 33.6 & & \textbf{34.4} & 20.0 & \textbf{32.0} & & 
           \textbf{29.3} & \textbf{16.0} & \textbf{27.4} & & 
           21.1 & \textbf{20.0} & 21.0 & & 
           18.0 & 8.0 & 16.9 & & 
           16.9 & 0.0 & 15.2 & & 
           \textbf{23.9} & 12.8 & \textbf{22.5}\\
\hline

\rotatebox[origin=c]{90}{GF} 
& TAIL
& 48.0 & & 48.8 & 40.0 & 47.3 & &
    48.0 & 28.0 & 45.1 & &
    48.0 & 68.0 & 50.5 & &
    46.5 & 16.0 & 43.1 & &
    44.4 & 0.0 & 40.0 & &
    47.1 & 30.4 & 45.2\Tstrut\\

\hline

& Seq-FT 
& 48.0 & & 0.0 & \textbf{84.0} & 14.0 & & 
           6.7 & 8.0 & 6.9 & & 
           1.1 & \textbf{68.0} & 9.5 & & 
           0.0 & 12.0 & 1.3 & & 
           2.2 & \textbf{12.0} & 3.2 & & 
           2.0 & \textbf{36.8} & 7.0\Tstrut\\
& EWC 
& 48.0 & & 8.0 & 56.0 & 16.0 & & 
           13.3 & \textbf{16.0} & 13.7 & & 
           1.1 & 44.0 & 6.5 & & 
           1.5 & 4.0 & 1.8 & & 
           2.7 & 0.0 & 2.4 & & 
           5.3 & 24.0 & 8.1\\
& TAIL-A 
& 48.0 & & 29.6 & 10.0 & 26.3 & & 
           20.7 & 6.7 & 18.7 & &
           16.1 & 6.0 & 14.9 & &
           12.0 & 3.2 & 11.0 & &
           11.7 & 0.0 & 10.5 & &
           18.0 & 5.2 & 16.3\\
\hline[dashed]
& TIB
& 48.0 & & 17.6 & 24.0 & 18.7 & & 
           16.0 & \textbf{16.0} & 16.0 & & 
           \textbf{20.0} & 56.0 & \textbf{24.5} & & 
           19.5 & \textbf{32.0} & 20.9 & & 
           13.3 & 0.0 & 12.0 & & 
           17.3 & 25.6 & 18.4 \Tstrut\\
& iManip 
& 48.0 & & 32.8 & 28.0 & 32.0 & & 
           20.7 & 0.0 & 17.7 & &
           17.7 & 24.0 & 18.5 & &
           21.0 & 16.0 & 20.4 & &
           18.7 & 0.0 & 16.8 & &
           22.2 & 13.6 & 21.1 \\
& GROW 
& \textbf{49.6} & & \textbf{38.4} & 32.0 & \textbf{37.3} & & 
           \textbf{31.3} & 8.0 & \textbf{28.0} & & 
           16.0 & 24.0 & 17.0 & & 
           \textbf{21.5} & 24.0 & \textbf{21.7} & & 
           \textbf{22.7} & 0.0 & \textbf{20.4} & & 
           \textbf{26.0} & 17.6 & \textbf{24.9}\\
\hline

\rotatebox[origin=c]{90}{MG}
& TAIL
& 60.0 & & 64.0 & 68.0 & 64.7 & &
    61.3 & 24.0 & 56.0 & &
    60.0 & 80.0 & 62.5 & &
    65.0 & 20.0 & 60.0 & &
    55.6 & 0.0 & 50.0 & &
    61.2 & 38.4 & 58.6\Tstrut\\

\hline

& Seq-FT 
& \textbf{60.0} & & 12.8 & \textbf{72.0} & 22.7 & & 
           3.3 & \textbf{40.0} & 8.6 & & 
           4.0 & \textbf{64.0} & 11.5 & &
           2.5 & 12.0 & 3.6 & & 
           2.7 & \textbf{12.0} & 3.6 & & 
           5.1 & \textbf{40.0} & 10.0
\Tstrut\\
& EWC 
& \textbf{60.0} & & 8.8 & 28.0 & 12.0 & & 
           11.3 & 8.0 & 10.9 & & 
           1.7 & 52.0 & 8.0 & & 
           5.5 & 4.0 & 5.3 & & 
           4.0 & 0.0 & 3.6 & & 
           6.3 & 18.4 & 8.0\\
& TAIL-A 
& \textbf{60.0} & & 34.0 & 16.0 & 31.0 & & 
           26.0 & 8.0 & 23.4 & & 
           16.6 & 10.0 & 15.8 & & 
           14.5 & 13.6 & 14.4 & & 
           10.8 & 0.7 & 9.8 & & 
           20.4 & 9.7 & 18.9\\
\hline[dashed]
& TIB
& \textbf{60.0} & & 16.0 & 28.0 & 18.0 & & 
           28.7 & 0.0 & 24.6 & & 
           19.4 & 56.0 & 24.0 & & 
           18.0 & 8.0 & 16.9 & & 
           25.3 & 0.0 & 22.8 & & 
           21.5 & 18.4 & 21.3 \Tstrut\\
& iManip 
& \textbf{60.0} & & 33.6 & 28.0 & 32.7 & & 
           24.0 & 16.0 & 22.9 & &
           16.0 & 24.0 & 17.0 & &
           22.0 & 12.0 & 20.9 & &
           16.0 & 0.0 & 14.4 & &
           22.3 & 16.0 & 21.6 \\
& GROW 
& 59.2 & & \textbf{40.0} & 36.0 & \textbf{39.3} & & 
           \textbf{41.3} & 16.0 & \textbf{37.7} & & 
           \textbf{25.1} & 28.0 & \textbf{25.5} & & 
           \textbf{33.5} & \textbf{16.0} & \textbf{31.6} & & 
           \textbf{30.2} & 0.0 & \textbf{27.2} & & 
           \textbf{34.0} & 19.2 & \textbf{32.3}\\
\hline
\end{tblr}
}
\label{tab:b5-5n1_performance}
\vspace{-5pt}
\end{table}

\begin{figure}[t]
\centering
\begin{minipage}[t]{0.6\textwidth}
\vspace{0pt}
\centering
\huge
\setlength{\tabcolsep}{4pt}
\captionsetup{skip=5pt}
\captionof{table}{\textcolor{black}{Ablation studies under B5-5N1. Here, figures denote performance averaged over all incremental stages for compactness. Stage-wise performance is in our appendix.}}
\vspace{3pt}
\resizebox{.9\textwidth}{!}{
\begin{tblr}{ 
        colspec             = {c | c | c | *{3}{c} | c |*{4}{c}},
        cell{1}{1,2} = {r=2}{c},
        cell{1}{3, 7}        = {c=4}{c}, 
        cell{13}{1-Z} = {bg=gray!25}
    }
\hline
ID & Ablation & Incremental Only & & & & Base + Incremental & & & \\
\hline
& & Base & Old & New & All & Base & Old & New & All \\
\hline
1 & w/o next grounding & 
59.2 & 27.9 & 13.6 & 26.1 &
47.2 & 19.0 & \textbf{29.6} & 20.4\\

2 & w/o goal grounding & 
59.2 & 32.9 & 20.8 & 31.5 &
52.0 & 22.1 & 13.6 & 21.1\\

3 & w/o action chain predictor & 
59.2 & 27.0 & \textbf{24.0} & 26.6 &
53.0 & 24.4 & 11.2 & 22.7\\

4 & predict next step only & 
59.2 & 7.8 & 17.6  & 9.3 &
58.4 & 28.1 & 7.2 & 25.3\\

5 & TIB + our LoRA-based update & 
59.2 & 21.9 & 10.4 & 20.4 &
- & - & - & -\\

6 & Chain herding vs. TIB herding & 
59.2 & 29.3 & 13.6 & 27.3 &
 - & - & - & - \\
 
\hline[dashed]

7 & w/o chain feat injection & 
59.2 & 31.3 & 15.2 & 29.4 &
54.4 & 24.5 & 18.4 & 23.8\Tstrut\\

8 & w/o chain decay weight & 
59.2 & 26.8 & 16.8 & 25.4 &
55.3 & 25.0 & 15.2 & 23.8\\

9 & w/o lang token in & 
59.2 & 24.3 & 15.2 & 23.2 &
48.8 & 28.7 & 9.6 & 26.1\\

\hline

- & iManip & 
\textbf{60.0} & 22.3 & 16.0 & 21.6 &
\textbf{60.0} & 22.3 & 16.0 & 21.6\Tstrut\\

- & GROW & 
59.2 & \textbf{34.0} & 19.2 & \textbf{32.3} & 
59.2 & \textbf{34.0} & 19.2 & \textbf{32.3}\\
\hline
\end{tblr}}
\label{tab:main_ablation}
\vspace{-5pt}
\end{minipage}
\hfill
\begin{minipage}[t]{0.37\textwidth}
    \vspace{0pt}
    \centering
    \includegraphics[width=\linewidth]{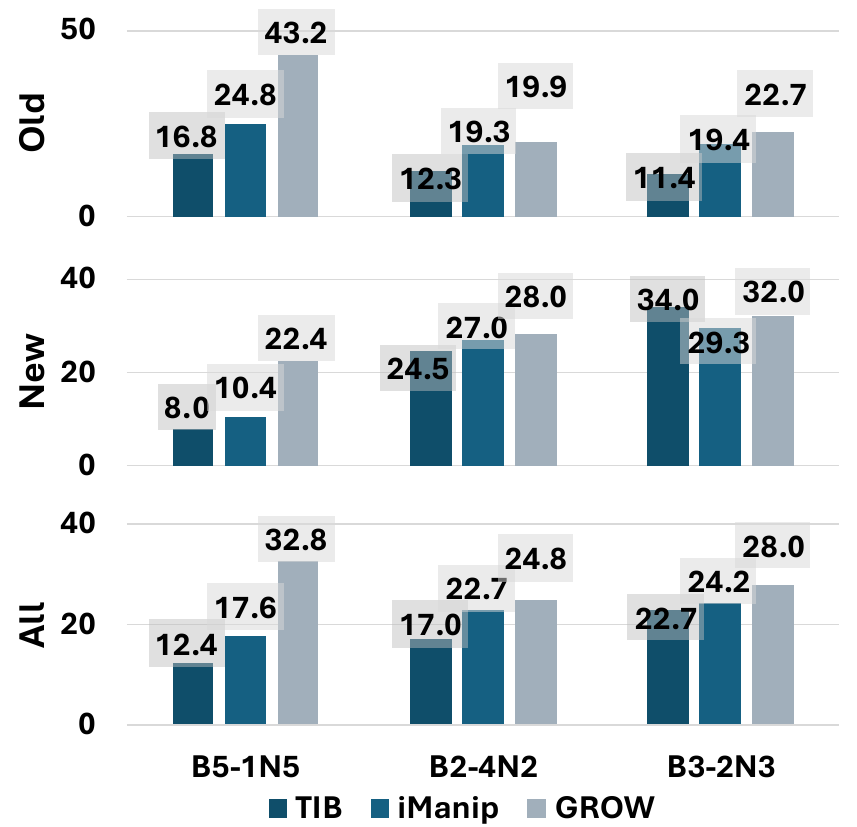}
    \captionsetup{skip=5pt}
    \captionof{figure}{Performance of B5-1N5, B2-2N4, and B3-2N3.}
    \label{fig:xSettings}
    \vspace{-5pt}
\end{minipage}
\vspace{-10pt}
\end{figure}

\subsection{Main Comparison with SOTA Methods}
As shown in Table~\ref{tab:b5-5n1_performance}, we compared GROW under the B5-5N1 setting using three different backbones, including PA, GF, and MG. Compared with baseline approaches, GROW consistently achieves SOTA performance across all backbones in terms of average overall performance. Specifically with MG, it outperforms iManip across old skills preservation, new skill comprehension, and overall performance, with a noticeable gap of $11.7\%$, $3.2\%$, and $10.7\%$, respectively. Similar improvement can be observed on GF and PA, where GROW surpasses iManip by $3.8\%$ and $3.5\%$ on overall performance, respectively. For new skill comprehension, Seq-FT and TIB achieve superior performance to GROW, mainly because they tune the global parameters of networks, which also causes their more severe degradation in old skill preservation. This demonstrates that GROW achieves a better balance between new-skill acquisition and old-skill retention.

\begin{table}[t]
\centering
\huge
\setlength{\tabcolsep}{4pt}
\caption{B5-1N1 performance.}
\vspace{-5pt}
\resizebox{.9\textwidth}{!}{
\begin{tblr}{ 
        colspec = {l | c *{25}{c}},
        cell{1}{1} = {r=2}{l},
        cell{1}{2,3,7,11,15,19,23} = {r=2}{c}, 
        cell{1}{4,8,12,16,20,24} = {c=3}{c},
        cell{5}{1-Z} = {bg=gray!25}
    }
\hline
Method & Base & 
& Slide the Block & & & 
& Put in Drawer & & & 
& Reach and Drag & & & 
& Push Buttons & & & 
& Stack Blocks & & & 
& Avg & & \\
\hline
& & 
& Old & New & All 
& & Old & New & All 
& & Old & New & All 
& & Old & New & All 
& & Old & New & All 
& & Old & New & All\Tstrut\\
\hline

TIB 
& \textbf{60.0} & 
& 16.0 & 28.0 & 18.0 
& & 16.8 & 0.0 & 14.0 
& & 14.4 & \textbf{52.0} & 20.7 
& & 17.6 & \textbf{48.0} & 22.7 
& & 16.8 & 4.0 & 14.7 
& & 16.3 & \textbf{26.4} & 18.0\Tstrut\\

iManip  
& \textbf{60.0} & 
& 33.6 & 28.0 & 32.7 
& & 28.8 & \textbf{12.0} & 26.0 
& & 24.8 & 12.0 & 22.7 
& & 18.4 & 16.0 & 18.0 
& & 23.2 & \textbf{8.0} & 20.7 
& & 25.8 & 15.2 & 24.0 \\

GROW
& 59.2 & 
& \textbf{40.0} & \textbf{36.0} & \textbf{39.3} 
& & \textbf{37.6} & \textbf{12.0} & \textbf{33.3} 
& & \textbf{43.2} & \textbf{52.0} & \textbf{44.7} 
& & \textbf{44.8} & 24.0 & \textbf{41.3} 
& & \textbf{28.0} & \textbf{8.0} & \textbf{24.7} 
& & \textbf{38.7} & \textbf{26.4} & \textbf{36.5} \\
\hline

\end{tblr}
}
\label{tab:b51n1_performance}
\end{table}

\begin{figure}[t]
    \centering
    \includegraphics[width=.9\linewidth]{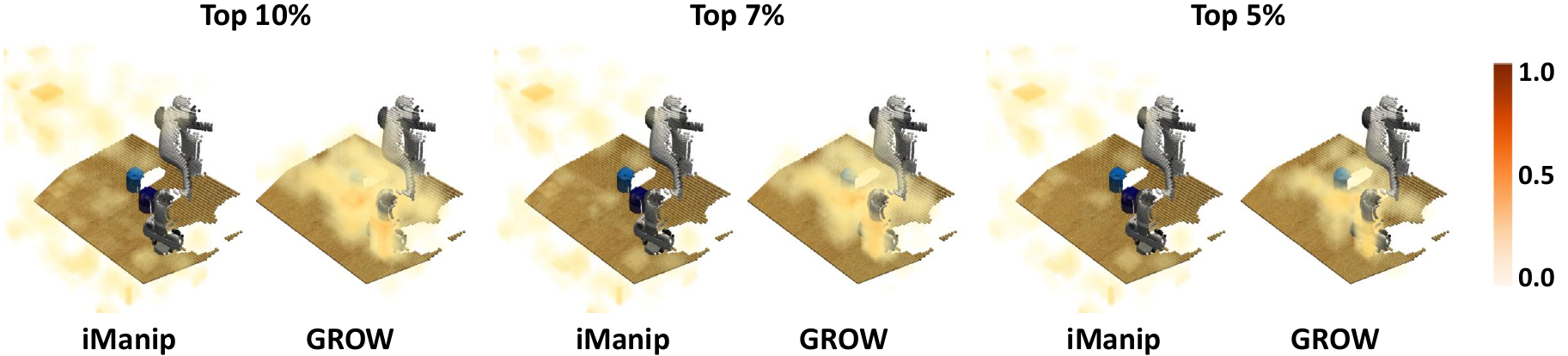}
    \vspace{-5pt}
    \caption{Visualization of attention heatmaps from iManip and GROW (Stage 1 of B5-5N1). Here we visualize the voxel-wise heatmaps with the top-$\mathrm{X}\%,\mathrm{X}\in[10, 7, 5]$ attention scores. }
    \label{fig:grounding_viz}
    \vspace{-5pt}
\end{figure}

\begin{table}[t]
\centering
\huge
\setlength{\tabcolsep}{4pt}
\caption{\textcolor{black}{Comparison of real-world rollouts. With 10 rollouts per evaluation, Wilson $95\%$ intervals span about $\pm30$ points.}}
\vspace{-5pt}
\resizebox{.9\textwidth}{!}{
\begin{tblr}{ 
        colspec = {c | *{11}{c} | *{3}{c}},
        cell{1}{1} = {r=2}{c},
        cell{1}{2,6,10,13} = {c=3}{c},
        cell{1}{5,9} = {r=2}{c}, 
        cell{2-Z}{4, 8, 12, 15} = {bg=gray!25}
    }
\hline
Manipulation Skill 
& Base & & &
& Stage 1 & & & 
& Stage 2 & &
& Avg & & \\
\hline
& Seq-FT & iManip & GROW 
& & Seq-FT & iManip & GROW 
& & Seq-FT & iManip & GROW
& Seq-FT & iManip & GROW\Tstrut\\
\hline
Put items in baskets 
& 20.0 & 20.0 & \textbf{60.0}
& & 10.0 & 10.0 & \textbf{40.0}
& & 0.0 & 10.0 & \textbf{30.0}
& 10.0 & 13.3 & \textbf{43.3}\\

Stack blocks 
& - & - & -
& & 30.0 & 0.0 & \textbf{50.0}
& & 0.0 & 10.0 & \textbf{30.0}
& 15.0 & 5.0 & \textbf{40.0} \\
Slide blocks to regions 
& - & - & -
& & - & - & -
& & 30.0 & 30.0 & \textbf{40.0}
& 30.0 & 30.0 & \textbf{40.0} \\
\hline

\end{tblr}
}
\label{tab:realworld_performance}
\vspace{-10pt}
\end{table}

\begin{figure}[t]
    \centering
    \includegraphics[width=\linewidth]{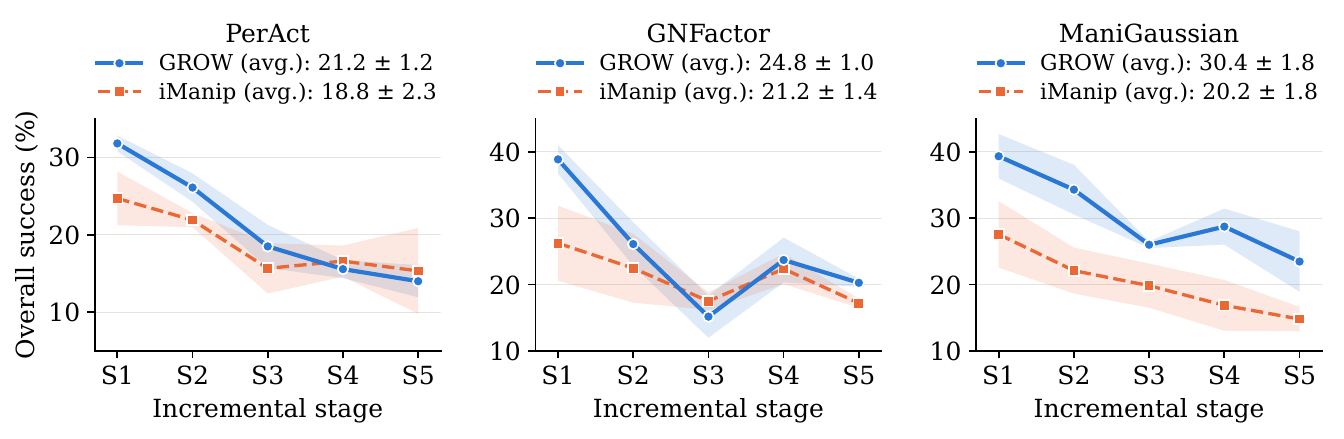}
    \caption{\textcolor{black}{Incremental-only uncertainty estimation conducted over three seeds sharing the same base checkpoint and data (mean ± std); y-axis ranges differ across backbones.}}
    \label{fig:ic_uncertainty}
\vspace{-10pt}
\end{figure}

\subsection{Ablation Studies}
\textbf{Component-wise Effectiveness.}
We first validate the contributions of the key components that compose GROW (Table~\ref{tab:main_ablation}) under two scenarios: disabling each component only during the incremental learning (\emph{Incremental Only}), and removing the component from both base pretraining and incremental stages (\emph{Base + Incremental}). Removing either next-keyframe or goal grounding consistently degrades the overall performance, confirming the complementary benefits of immediate interaction and long-horizon goal supervision. \textcolor{black}{Removing the chain predictor drops overall success from $32.3\%$ to $26.6\%$, and restricting it to the next keyframe from scratch lowers the overall success from $34.0\%$ to $28.1\%$ (Appendix~\ref{a_sec:stage_wise_ablation}), emphasizing the importance of modeling skill progression.} Replacing chain-based herding with keyframe-based herding also causes a noticeable drop from $32.3\%$ to $27.3\%$, illustrating the improvement brought by long-horizon-aware replay selection. Adopting GROW's identical LoRA-based update for TIB (ID~5) yields $20.4\%$ average overall success, on par with the original TIB ($21.3\%$) and 11.9 points below GROW, confirming that the gains stem from GROW's components rather than its update mechanism.

\textbf{Specific Designs of Action Chain Predictor.}
We further evaluate some key designs in the action chain predictor (Table~\ref{tab:main_ablation}). Without the aid of chain feature injection, both incremental and full training suffer from a non-trivial performance decrease, indicating the importance of using long-horizon representation for next-step prediction. Disabling the decay weight causes a drop of more than $7\%$ in old skill preservation, suggesting that emphasizing nearer-term progression while retaining long-horizon context improves old-skill preservation. Removing the linguistic features from input leads to a similar gap, supporting the importance of explicit instruction conditioning of long-horizon planning.

\textbf{Performance under Diverse SIL Settings.}
To rule out the setting bias, we evaluate GROW under 4 more SIL settings along with the competitive TIB and iManip. As in Table~\ref{tab:b51n1_performance}, the improvement brought by GROW sustains across all short-term SIL with varying new tasks, outperforming existing methods by more than $10\%$ on average. Similar improvements can also be observed on B5-1N5, B2-4N2, and B3-2N3 (Figure~\ref{fig:xSettings}) across almost all metrics and settings.

\textcolor{black}{\textbf{Uncertainty Estimation.}
To assess the reliability of our comparison, we repeat the incremental stages of B5-5N1 under three random seeds for each backbone, keeping the base checkpoint, training data, and evaluation scenes fixed. As shown in Figure~\ref{fig:ic_uncertainty}, GROW improves the stage-averaged overall success over iManip on all three backbones, consistently on ManiGaussian and GNFactor, while the margin on PerAct is comparable to the seed variation and concentrated in the early stages. Part of this variation stems from stochastic evaluation, as re-evaluating the same base checkpoint yields 62.4 $\pm$ 1.8\%. Fully independent runs with regenerated data and retrained base policies further confirm consistent gains on PerAct and GNFactor (Appendix~\ref{a_sec:uncertainty}).}

\textbf{Qualitative Comparison.}
We present the heatmap comparison of the early $\texttt{CrossAttn}_\textrm{enc}$ between iManip and GROW in Figure~\ref{fig:grounding_viz}. By incorporating the dual-horizon grounding, GROW can achieve much better concentration during the early perception stage, with a main focus on the task-relevant regions (i.e., jars, the gripper, and workspace). 

\subsection{Real-World Rollouts}
For the real-world experiments, we compare GROW with Seq-FT and iManip using the backbone PerAct~\citep{shridhar2023perceiver} under a SIL setting of B1-2N1. Throughout the SIL process, GROW achieves the highest success rate across all manipulation tasks compared with Seq-FT and iManip. Note that GROW also outperforms the baselines in the base task ``Put items in baskets'', mainly because our task setting involves localizing and picking small objects (e.g., a strawberry), where the perception grounding of GROW could help substantially. Details are elaborated in our appendix.

\textcolor{black}{\subsection{Limitations}
\label{sec:limitations}
GROW has several limitations. First, its long-horizon cues are positional: the goal anchor is the terminal end-effector position, and the action chain predicts only the translations within a finite horizon ($k=7$). Skills whose progression depends critically on rotation, or extends beyond this horizon (e.g., ``stack blocks''with 21 keyframes on average (Table~\ref{a_tab:keyframe-statistics}), where all methods remain low), are therefore not fully captured. Second, dual-horizon grounding operates on encoder cross-attention over voxel tokens, which restricts GROW to PerceiverIO-based keyframe policies, while its applicability to other architectures remains unexplored. Third, chain-based herding groups replay candidates by keyframe index, which does not always correspond to the same execution stage across demonstrations of different lengths; a progress-aware grouping may further improve replay selection. Finally, our real-world study is limited in scale (three skills, 10 rollouts per evaluation, a single run) and serves as a demonstration of applicability rather than statistical evidence, which rests instead on our simulation experiments.}

\section{Conclusion}
In this work, we study skill-incremental robotics manipulation from the perspective of the inherent level mismatch between local keyframe-wise learning and long-horizon semantics required to preserve the skill-level knowledge. To address this gap, we propose GROW, which incorporates long-term awareness throughout perception, action prediction, and replay-herding. Extensive experiments conducted on RLBench demonstrate the consistent improvement across different incremental settings and backbones, while ablation studies further justify the effectiveness of the proposed design. The supplemented real-world rollouts further demonstrate the applicability of GROW in real-life scenarios. Our results highlight the significance of long-horizon semantic preservation for continual robot learning, paving the road for future extensions toward more complex skill compositions and long-term real-world adaptation.

\subsection*{AI use statement}

In this work, we used generative AI tools for implementing methods and designing or providing feedback on research methodology or experiments.
We have not used generative AI tools for generating synthetic data sets, proposing or refining hypotheses, assisting with translation, cleaning and reformatting datasets, supporting qualitative and thematic data analysis, or interpreting results,
and the rest of the required disclosure tasks do not apply to this work.
Additionally, we used generative AI tools for creating or editing software code and editing a research paper to improve readability. We have reviewed all AI-assisted work. LLM-generated code was verified and tested for correctness through multiple steps of verification with intermediate results and inspection of the authors. The final manuscript is proofread by the authors. We take responsibility for the final content of this work, including text, claims, or artifacts produced with the aid of generative AI.

\subsection*{Reproducibility statement}
Code of algorithms, simulation benchmarking, and real-world rollouts are included in our supplementary materials and will be released upon acceptance. Details of simulation implementation and real-world rollout protocols are thoroughly introduced in the appendix. For any datasets used in
the experiments, a complete description of the data processing is provided in the appendix.

\bibliography{references}
\bibliographystyle{iclr2027_conference}

\appendix
\clearpage

\section*{Appendix Contents}
In our appendix, we elaborate on the details of the following aspects:

\begin{itemize}
    \item A. Implementation Details
    \item B. Additional Experimental Results
    \item C. Hyperparameter Analysis
    \item D. Real-World Setup
\end{itemize}

\section{Implementation Details}
In this section, we detail the benchmark details and method implementation details. For benchmark details, we elaborate on the structure of each incremental setting, skill definition, and variations. In terms of method implementation details, we first introduce the universal settings that apply to all methods, and then provide the method-specific modifications. Code can be found in our supplementary materials.

\subsection{Benchmark Details}
As mentioned in Section~\ref{sec:exp}, we follow the benchmark protocols of iManip~\citep{zheng2025imanip} to re-implement the incremental benchmark. For visual observation, we adopt RGB-D images from a single front camera with a resolution of $128\times128$ as the universal input for all backbones. The RGB-D images are then projected as colorful 3D voxels, which are further encoded by a 3D convolutional network before input into the PercieverIO. Specifically for GNFactor and ManiGaussian, which contain 3D rendering modules that require multi-view images as supervision signals, the same setting of 20 cameras is used specifically for their rendering modules. Linguistic instructions are encoded by CLIP RN50~\citep{radford2021learning}, while the proprioception data is encoded by an MLP layer. Following the backbone settings~\citep{shridhar2023perceiver,ze2023gnfactor,lu2024manigaussian}, we utilize the same discrete presentation for position and rotation formatting, where $\mathbf{a}^\mathrm{pos}_t\in\mathbb{R}^{100^3}$ is a one-hot matrix that represents a $1.0\mathrm{m}^3$ volume indicating the voxel where the end-effector position would approach and $\mathbf{a}^\mathrm{rot}_t\in\mathbb{R}^{(360/5)\times3}$ denotes one-hot rotation GT with a discretization of $5^\circ$. During training and evaluation, the workspace in simulation is bounded from \([-0.3,-0.5,0.6]\) to \([0.7,0.5,1.6]\) m (x, y, z).

\begin{table}[!ht]
\centering
\caption{Task variations and train/test coverage.}
\label{a_tab:task_variations}
\vspace{-5pt}
\resizebox{.9\textwidth}{!}{
\begin{tblr}{
    colspec = {c l c X[l]},
    row{1} = {font=\bfseries},
    hline{1,Z} = {1pt},
    hline{2} = {0.6pt},
    rows = {valign=m},
    columns = {halign=c},
    column{2} = {halign=l},
    column{4} = {halign=l},
}
ID & Skill &
Variations &
Variation semantics \\

1 
& \texttt{close\_jar}
& 20
& Target jar color\\

2
& \texttt{open\_drawer}
& 3
& Bottom, middle, or top drawer\\

3
& \texttt{sweep\_to\_dustpan\_of\_size}
& 2
& Tall or short dustpan\\

4
& \texttt{meat\_off\_grill}
& 2
& Chicken or steak\\

5
& \texttt{turn\_tap}
& 2
& Left or right tap\\

6
& \texttt{slide\_block\_to\_color\_target}
& 4
& Green, blue, pink, or yellow target\\

7
& \texttt{put\_item\_in\_drawer}
& 3
& Bottom, middle, or top drawer\\

8
& \texttt{reach\_and\_drag}
& 20
& Target region color\\

9
& \texttt{push\_buttons}
& 50
& Ordered colored-button sequence, requiring 1--3 presses\\

10
& \texttt{stack\_blocks}
& 60
& 20 target colors $\times$ stack heights of 2, 3, or 4\\
\end{tblr}}
\end{table}

\begin{table}[t]
    \centering
    \caption{\textcolor{black}{Statistics of discovered keyframe counts in training data per task.}}
    \label{a_tab:keyframe-statistics}
    \begin{tblr}{
        colspec = {Q[l] Q[r] Q[r] Q[r] Q[r]},
        row{1} = {font=\bfseries}
    }
        \toprule
        Task & Episodes & Mean $\pm$ Std. & Min & Max \\
        \midrule
        \texttt{close\_jar}
            & 20 & $6.950 \pm 0.669$ & 6 & 8 \\
        \texttt{meat\_off\_grill}
            & 20 & $6.550 \pm 0.669$ & 5 & 7 \\
        \texttt{open\_drawer}
            & 20 & $4.350 \pm 0.654$ & 3 & 5 \\
        \texttt{push\_buttons}
            & 20 & $5.400 \pm 1.985$ & 2 & 8 \\
        \texttt{put\_item\_in\_drawer}
            & 20 & $14.650 \pm 1.276$ & 12 & 16 \\
        \texttt{reach\_and\_drag}
            & 20 & $6.550 \pm 0.589$ & 6 & 8 \\
        \texttt{slide\_block\_to\_color\_target}
            & 20 & $4.550 \pm 1.465$ & 2 & 7 \\
        \texttt{stack\_blocks}
            & 20 & $21.250 \pm 5.726$ & 12 & 29 \\
        \texttt{sweep\_to\_dustpan\_of\_size}
            & 20 & $6.350 \pm 0.654$ & 5 & 7 \\
        \texttt{turn\_tap}
            & 20 & $3.050 \pm 0.740$ & 2 & 4 \\
        \bottomrule
    \end{tblr}
\end{table}

The incremental benchmark is composed of 10 tasks listed in Table~\ref{a_tab:task_variations}. Under each incremental setting of B\textit{n}-\textit{k}N\textit{m}, the skills of ID 1-\textit{n} are utilized for base pre-training, while the following stages sequentially take \textit{m} tasks per stage. For all runs, we generate the training demonstrations and define evaluation rollouts using the same random seed (seed=0) for fair comparison. Most incremental settings follow the same iteration protocols, with 100,000 iterations allocated for base pre-training and 50,000 used for incremental learning per stage. The only exception is B2-4N2, where we reduce the base iterations to 50,000 since it involves much fewer base tasks during the pre-training stage. While iManip~\citep{zheng2025imanip} mentions that it utilizes 100,000 iterations for training, it does not explicitly point out whether this is for base pre-training or incremental update. Our 50,000-step setting is therefore deduced from its ablation study in Table 3 and our empirical estimation based on the number of tasks for each stage.

RLBench skills with varying task configurations. \texttt{open\_drawer} and \texttt{put\_item\_in\_drawer} each contain three variations corresponding to the bottom, middle, and top drawers; \texttt{sweep\_to\_dustpan\_of\_size}, \texttt{meat\_off\_grill}, and \texttt{turn\_tap} each contain two variations, respectively specifying the dustpan size, meat type, and tap side; and \texttt{slide\_block\_to\_color\_target} contains four color-target variations. The more diverse tasks include \texttt{close\_jar} and \texttt{reach\_and\_drag}, each with 20 target-color variations; \texttt{push\_buttons}, with 50 variations combining ordered colored-button sequences and one to three required presses; and \texttt{stack\_blocks}, with 60 variations combining 20 target colors with stack heights of two, three, or four blocks. In total, these tasks span 168 predefined skill variations. Each episode samples from the corresponding task’s full variation space, although the finite training and evaluation sets do not necessarily contain every variation for high-cardinality tasks. \textcolor{black}{We further supplement the statistics of keyframe numbers of our training data in Table~\ref{a_tab:keyframe-statistics}.}

\subsection{Method Implementation Details}
In this section, we first detail the specific design of GROW and our hyperparameter selection. Subsequently, we introduce the implementation of all baselines.

\begin{table*}[t]
\centering
\caption{Per-layer architecture of the chain predictor.}
\label{a_tab:chain_predictor_architecture}
\begin{tblr}{
    width=\textwidth,
    colspec={
        Q[l,m,wd=0.19\textwidth]
        Q[l,m,wd=0.24\textwidth]
        Q[l,m,wd=0.48\textwidth]
    },
    row{1}={font=\bfseries},
    cells={valign=m},
}
\toprule
Layer / Component
    & Input $\rightarrow$ Output
    & Function \\
\midrule

Perceiver latent pooling
    & $2048 \times 512 \rightarrow 1 \times 512$
    & Mean-pools the final Perceiver latent tokens into one scene token. \\

Language projection
    & $77 \times 128 \rightarrow 77 \times 512$
    & Projects the preprocessed CLIP token sequence to the chain-feature dimension. \\

Scene-language context
    & $(1+77) \times 512$
    & Concatenates the pooled scene token and projected language tokens. \\

Keyframe queries
    & $7 \times 512$
    & Uses seven ordered learnable query slots; valid leading slots decode consecutive future actions, with the last valid slot assigned to the goal. \\

Cross-attention
    & $7 \times 512 \rightarrow 7 \times 512$
    & Uses four-head attention to extract information from the 78 scene-language tokens. \\

Cross-attention LayerNorm
    & $7 \times 512 \rightarrow 7 \times 512$
    & Applies residual addition followed by feature normalization. \\

Causal self-attention
    & $7 \times 512 \rightarrow 7 \times 512$
    & Models temporal dependencies while restricting each query to itself and earlier slots. \\

Self-attention LayerNorm
    & $7 \times 512 \rightarrow 7 \times 512$
    & Applies residual addition followed by feature normalization. \\

Translation head
    & $512 \rightarrow 3 \times 100$
    & Predicts categorical voxel distributions independently for the $x$, $y$, and $z$ axes. \\

Chain output
    & $7 \times 3 \times 100$
    & Represents the translation logits for all predicted future keyframes. \\

Chain-context pooling
    & $7 \times 512 \rightarrow 1 \times 512$
    & Pools detached query features using normalized weights proportional to $0.7^i$. \\

Decoder injection
    & $2048 \rightarrow 2049$ context tokens
    & Appends the pooled chain token to the original decoder cross-attention context. \\

\bottomrule
\end{tblr}
\end{table*}

\subsubsection{Implementation Details of GROW}
\textbf{Hyperparameter setting.} GROW introduces three major components to each keyframe-based backbone during both base pre-training and incremental stages, where most of the specifications remain identical. For dual-horizon grounding, both Gaussian anchors $\mathbf{g}^{\textrm{next}}_t$ and $\mathbf{g}^{\textrm{goal}}_t$ share the same configuration with $\sigma=3$ (Eq.~\ref{Eq:ground_target}). The corresponding coefficient $\lambda_\mathrm{n}$ and $\lambda_\mathrm{g}$ are both set as $0.01$ (Eq.~\ref{Eq:grounding_loss}). As for the action chain predictor, the length of learnable queries $\mathcal{Q}_\mathrm{chain}$ is set to $k=7$, while the decay factor $\tau$ is set to $0.7$ (Eq.~\ref{Eq:chain_xA}, Eq.~\ref{Eq:chain_loss}, Eq.~\ref{Eq:chain_feat}, and Eq.~\ref{Eq:lt_descriptoer}). In terms of chain-based replay herding, we allocate a budget of $B_m=2$ for each previous skill per average keyframe (Eq.~\ref{Eq:chain_replay}). The loss coefficient $\lambda_\mathrm{dis}$ is 0.01 while $\lambda_\mathrm{chain}$ is set as 0.001. We conduct a series of sensitivity analyses across some important hyperparameters.

\textbf{Architecture of Action Chain Predictor.} 
We list the per-layer components of the action chain predictor in Table~\ref{a_tab:chain_predictor_architecture} along with their corresponding function. Overall speaking, the proposed action chain predictor is an intuitive PerceiverIO-similar module, using learnable queries to decode the action chain features. It is also a lightweight component, containing about 2.3M trainable parameters, which additionally occupy 35.5MB GPU memory; including activations, peak training memory is ~31.7 GiB and the median step time is 0.644 s, vs. 31.7 GiB / 0.630 s for iManip) during the training process. Despite being simple and intuitive, enabling the action chain predictor alone can lead to substantial improvement when using our GROW or TIB baseline, as we will show in the following section.

\textbf{Details of LoRA-based Update.}
Here we elaborate on the LoRA-based update we used for GROW. At each incremental stage, we initialize a fresh rank-8 LoRA adapter (\(\alpha=16\)) on all PerceiverIO encoder cross-attention ($\texttt{CrossAttn}_\mathrm{enc}$ in Eq.~\ref{Eq:enc_crossattn}), latent self-attention ($\texttt{SelfAttn}_{\times M}$ in Eq.~\ref{Eq:sa_features}), and decoder cross-attention projections ($\texttt{CrossAttn}_\mathrm{dec}$ in Eq.~\ref{Eq:sa_features}). The voxel encoder and original PerceiverIO parameters are frozen, including their batch-normalization statistics, while the LoRA parameters, action decoder, chain predictor, and optional neural renderer from ManiGaussian and GNFactor are trained for 50,000 iterations. The learned LoRA updates are then merged into the corresponding attention weights via summation, after which only the merged attention projections are re-calibrated for 2,000 iterations using a \(0.1\times\) learning rate. During the re-calibration stage, the decoder and chain predictor remain frozen, and rendering and distillation are disabled. The merged and recalibrated model becomes the initialization for the next stage, where a new zero-delta LoRA adapter is introduced. In this way, we can maintain a finite size of trainable parameters while incrementally learning new skills.

\subsubsection{Implementation Details of Baselines}
We first introduce the universal setting shared across all baselines to make a fair comparison with GROW, and then detail the baseline-specific settings individually.

\textbf{Universal Setting.}
For all baselines, we implement them based on their original paper and adopt the same incremental training setting. Specifically, following the training protocols used for PerceverIO-based backbones~\citep{ze2023gnfactor,lu2024manigaussian}, all runs, including GROW, are trained using the LAMB optimizer~\citep{you2019large} with an initial learning rate of $5\times 10^{-4}$. Experiments are conducted with a batch size of 1, using a single NVIDIA RTX PRO 6000 GPU. Below are the baseline-specific configurations. For each backbone, all incremental training starts with the same base checkpoint (except for GROW, as it introduces additional modules during the base training) to make a fair comparison. Training demonstrations and evaluation scenes are generated with a fixed seed; rollouts remain stochastic, and seed variance is reported in Appendix~\ref{a_sec:uncertainty}.

\begin{itemize}
    \item \textbf{Seq-FT.} This is the intuitive sequential fine-tuning using each backbone. For each incremental stage, the backbone is fine-tuned using the optimizer and checkpoint inherited from the end of the previous stage. Only the training data from the unseen skills are observable, and no replays are involved for each incremental stage. 
    
    \item \textbf{EWC.} We implement EWC using the canonical Online-EWC formulation~\citep{kirkpatrick2017overcoming} without replaying old-task demonstrations. Each incremental stage is trained only on the newly introduced task while all BC-policy parameters remain trainable. After each stage, an empirical diagonal Fisher matrix is estimated from the squared gradients of the action-only behavioral-cloning loss over random 1,000 valid transitions from that stage. The Fisher statistics are accumulated as \(\bar{F}_{k}=\gamma\bar{F}_{k-1}+F_k\), with \(\gamma=0.9\), and the final policy parameters are stored as the new anchor. At the next stage, deviations from this anchor are penalized using a Fisher-weighted quadratic regularizer with \(\lambda_{\mathrm{EWC}}=100\). The protected scope covers the active voxel encoder, language and proprioceptive processing, Perceiver transformer, and action decoder, while excluding the frozen language model and neural renderer. Batch-normalization statistics are frozen during incremental training.

    \item \textbf{TAIL-A.} Our task-agnostic TAIL implementation uses the jointly trained base policy as an immutable shared backbone. For every incremental task, an independent expert is initialized directly from this base policy, rather than from the preceding incremental expert. Each expert trains a fresh rank-8 LoRA adapter with $\alpha=16$ on all PerceiverIO attention projections together with a complete task-specific action decoder, while the remaining backbone parameters are frozen. TAIL uses only behavioral cloning on the current task. At evaluation, no task identity is used to select an expert. The primary task-agnostic result uniformly averages the success rates obtained by the stage-0 policy and all available task experts, while an additional probability-ensemble variant averages their normalized translation, rotation, gripper, and collision distributions at every action step.

    \item \textbf{TIB.} TIB is a keyframe-based replay baseline introduced in iManip~\citep{zheng2025imanip}, which retains a compact set of representative old-task keyframe transitions and combines exemplar replay with output distillation. After each stage, every transition is represented by a 220-dimensional action feature formed by concatenating the predicted rotation/gripper probabilities and collision probabilities. Transitions are grouped according to their ordinal keyframe position, L2-normalized, and selected independently within each group using greedy mean herding. We retain 2 transitions per keyframe position by default to form a fair comparison with GROW using the same amount of replay budget (i.e., $B_m=2$). During the following stage, all new-task transitions are trained jointly with the selected old-task exemplars using task-uniform replay. A frozen copy of the preceding policy provides an old-task distillation target, with an MSE loss between the old and current translation, rotation, gripper, and collision probabilities weighted by \(\lambda_{\mathrm{dis}}=0.01\).

    \item \textbf{iManip.} iManip~\citep{zheng2025imanip} expands the Perceiver architecture at every incremental stage while freezing previously learned components. Following their protocol in the manuscript, for each stage, we introduce 16 new learnable prompt tokens on the input/context side of the encoder cross-attention and append a \(512\times8\) query extension and a \(512\times8\) key extension to every latent self-attention layer. With eight attention heads, this increases the per-head query/key dimension by one at each stage, while leaving the value projections, output projections, and \(2048\times512\) latent array unchanged. Previously learned prompts and query/key extensions are frozen, and only the newly added parameters and action decoder are optimized. Old-task memory is constructed using farthest-distance entropy sampling: transitions are grouped by keyframe position, scored by their total action cross-entropy over translation, rotation, gripper, and collision predictions, and selected to cover diverse uncertainty levels. Incremental training combines these exemplars with the complete new-task data and applies old-task action-probability distillation with weight \(\lambda_{\mathrm{dis}}=0.01\). 

\end{itemize}

\paragraph{Reproduction Details and Performance Gap.}
We observe that the reproduced iManip performance is lower than that reported in the original paper despite following its description. Several implementation details make an exact reproduction difficult. First, the published iManip model is described as a standalone Extendable PerceiverIO architecture, while an exact layer-wise mapping between this model and the PerAct-/ManiGaussian-style backbones used in its comparisons is not fully specified. We therefore implement the reported incremental mechanisms consistently on top of each backbone in our evaluation rather than reproducing an architecture specific to iManip.

Second, the original paper reports a batch size of 1, learning rate of $2\times10^{-3}$, and 100k training iterations~\citep{zheng2025imanip}. Its freezing analysis further explicitly trains each B5-1N1 experiment for 100k iterations and reports convergence at approximately 60k iterations for the selected configuration. However, the exact per-stage optimization schedule for the full B5-5N1 sequence is not separately documented. For a controlled comparison, we therefore adopt a fast tryout run on stage 1 of B5-5N1, using the same stage-wise training schedule for all incremental methods in our experiments. Importantly, directly applying the reported learning rate $2\times10^{-3}$ to our iManip implementation results in substantially lower performance (e.g., $9.3\%$ overall success at Stage~1), whereas reducing it to $5\times10^{-4}$ (the default learning rate shared across all PerceiverIO-based backbones) improves the corresponding performance to $32.7\%$ (Table~\ref{a_tab:imanip_im}). We consequently report the latter, stronger configuration for iManip throughout our experiments. 

\textcolor{black}{To verify whether iManip converges under the lower learning rate ($5\times10^{-4}$) adopted in our experiments, we conduct two additional runs: a full B5-5N1 run with 60k iterations per incremental stage, following the convergence point reported in Table~3 of \citet{zheng2025imanip}, and a verification run under B5-2N1 with 100k iterations per incremental stage, matching its reported training length. All iManip runs share the same base checkpoint and the same training and evaluation data. As shown in Table~\ref{a_tab:imanip_convergence}, increasing the number of incremental iterations does not improve iManip. With 60k iterations, the average overall success over the full sequence is $19.3\%$, compared with $21.6\%$ at 50k. Extending training to 100k iterations yields $21.3\%$ and $25.1\%$ overall success at Stages~1 and~2, respectively, again without a consistent gain. In contrast, GROW, trained with the default 50k schedule, achieves the highest overall success at every stage among all compared runs, exceeding the best iManip schedule by 10.7 points on average ($32.3\%$ vs. $21.6\%$). These results indicate that our 50k schedule does not under-train iManip, and that its gap to GROW cannot be attributed to insufficient incremental training.}

\begin{table}[t]
\centering
\huge
\setlength{\tabcolsep}{4pt}
\caption{Performance of iManip using different learning rates.}
\vspace{-5pt}
\resizebox{.4\textwidth}{!}{
\begin{tblr}{ 
        colspec = {c | c | *{3}{c}},
        cell{1}{1,2} = {r=2}{c},
        cell{1}{3} = {c=3}{c}
    }
\hline
Learning Rate & Base & Stage 1 \\
& & Old & New & All \\
\hline
0.002 & 60.0 & 8.0 & 16.0 & 9.3 \\
0.0005 & 60.0 & 33.6 & 28.0 & 32.7\\
\hline

\end{tblr}}\label{a_tab:imanip_im}
\end{table}

\begin{table}[t]
\centering
\huge
\setlength{\tabcolsep}{4pt}
\caption{\textcolor{black}{Convergence analysis of iManip. All runs of iManip use the same base checkpoint with identical training and testing data.}}
\vspace{-3pt}
\resizebox{\textwidth}{!}{
\begin{tblr}{ 
        colspec             = {l  c *{25}{c}},
        cell{1}{1} = {r=2}{l},
        cell{1}{2, 6, 10, 14, 18, 22}        = {r=2}{c}, 
        cell{1}{3, 7, 11, 15, 19, 23} = {c=3}{c},
        cell{5,6}{1-Z} = {bg=gray!25}
    }
\hline
Variants & & Stage 1 & & & & Stage 2 & & & & Stage 3 & & & & Stage 4 & & & & Stage 5 & & & & Avg \\
\hline
& & Old & New & All & & Old & New & All & & Old & New & All 
   & & Old & New & All & & Old & New & All & & Old & New & All\Tstrut\\
\hline

iManip (60k) 
& & 24.8 & 20.0 & 24.0 & & 
    24.0 & 0.0 & 20.6 & & 
    17.1 & 20.0 & 17.5 & &
    21.0 & \textbf{28.0} & 21.8 & &
    14.2 & 0.0 & 12.8 & &
    20.2 & 13.6 & 19.3 \\

iManip (100k) 
& & 21.6 & 20.0 & 21.3 & &
    28.0 & 8.0 & 25.1 & &
    - & - & - & &
    - & - & - & & 
    - & - & - & &
    - & - & - \\

iManip (50k)
& & 33.6 & 28.0 & 32.7 & & 
   24.0 & \textbf{16.0} & 22.9 & &
   16.0 & 24.0 & 17.0 & &
   22.0 & 12.0 & 20.9 & &
   16.0 & 0.0 & 14.4 & &
   22.3 & 16.0 & 21.6 \\

\hline

GROW (50k)
& & \textbf{40.0} & \textbf{36.0} & \textbf{39.3} & & 
           \textbf{41.3} & \textbf{16.0} & \textbf{37.7} & & 
           \textbf{25.1} & \textbf{28.0} & \textbf{25.5} & & 
           \textbf{33.5} & 16.0 & \textbf{31.6} & & 
           \textbf{30.2} & 0.0 & \textbf{27.2} & & 
           \textbf{34.0} & \textbf{19.2} & \textbf{32.3}\Tstrut\\

\hline
\end{tblr}
}
\label{a_tab:imanip_convergence}
\vspace{-5pt}
\end{table}

Finally, we retain the original farthest-distance entropy replay strategy and its budget of two samples per keyframe position. With such a limited budget, this diversity-oriented criterion can be sensitive to the distribution of action-prediction uncertainty, since only a small number of widely separated samples are retained. Nevertheless, we preserve the original selection rule rather than replacing it with an alternative herding strategy. Therefore, our iManip results should be interpreted as a controlled reimplementation of its reported incremental mechanisms under our common backbone and training protocol.

\section{Additional Experimental Results}
In this section, we first introduce some additional quantitative results to better justify the effectiveness of GROW, and then include more qualitative comparisons.

\subsection{Quantitative Results}

\begin{table}[t]
\vspace{0pt}
\centering
\caption{Baseline TIB with GROW's components. Here * indicates disabling the normalization statistics freezing. ``Chain'' represents the chain predictor. All runs tune only the action decoder during the incremental stages.}
\resizebox{.6\textwidth}{!}{
\begin{tabular}{l|c|cccc}
\hline
\multirow{2}{*}{Method} & \multirow{2}{*}{Steps} & \multirow{2}{*}{Base} & \multicolumn{3}{c}{Stage 1}\Tstrut\\
\cline{4-6}
& & & Old & New & All \Tstrut\\
\hline
MG-A & 50k+10k & 46.4 & 6.4 & 16.0 & 8.0 \\
MG-N & 50k+10k & 46.4 & 12.8 & 4.0 & 11.3 \\
MG-TIB* & 50k+10k & 46.4 & 3.2 & 12.0 & 4.7 \\
MG-TIB & 50k+10k & 46.4 & 24.8 & 0.0 & 20.7\\
MG-TIB & 50k+50k & 46.4 & 13.6 & 8.0 & 11.3\\
\hline
MG-TIB+$\mathcal{L}_\mathrm{dh}$ & 50k+10k & 44.8 & 34.4 & 24.0 & 32.7\Tstrut\\
MG-TIB+Chain & 50k+10k & 44.8 & 30.4 & 8.0 & 26.7 \\
MG-TIB+$\mathcal{L}_\mathrm{dh}$+Chain & 50k+10k & 45.6 & 34.4 & \textbf{28.0} & \textbf{33.3}\\
\hline
MG-TIB+$\mathcal{L}_\mathrm{dh}$ & 50k+50k & 44.8 & 16.0 & 20.0 & 16.7\Tstrut\\
MG-TIB+Chain & 50k+50k & 36.8 & 24.0 & \textbf{28.0} & 24.7 \\
MG-TIB+$\mathcal{L}_\mathrm{dh}$+Chain & 50k+50k & 42.4 & \textbf{35.2} & 16.0 & 32.0\\
\hline

\end{tabular}
}
\label{a_tab:tib+grow}
\end{table}

\begin{table*}[t]
\centering
\caption{GROW and TIB with different incremental update protocols. Here ``F'' indicates freezing encoder and PercieverIO's components and only update the action decoder, while ``L'' denotes the universal LoRA during incremental stages.}
\resizebox{\textwidth}{!}{
\begin{tabular}{l c ccc ccc ccc ccc ccc ccc}
\toprule
\multirow{2}{*}{Method} & \multirow{2}{*}{Base} 
  & \multicolumn{3}{c}{Stage 1} 
  & \multicolumn{3}{c}{Stage 2} 
  & \multicolumn{3}{c}{Stage 3} 
  & \multicolumn{3}{c}{Stage 4} 
  & \multicolumn{3}{c}{Stage 5} 
  & \multicolumn{3}{c}{Avg} \\
\cmidrule(lr){3-5}\cmidrule(lr){6-8}\cmidrule(lr){9-11}
\cmidrule(lr){12-14}\cmidrule(lr){15-17}\cmidrule(lr){18-20}
 & & Old & New & All & Old & New & All & Old & New & All 
   & Old & New & All & Old & New & All & Old & New & All \\
\midrule
TIB-F 
& \textbf{60.0} & 28.0 & 8.0 & 24.7 & 21.3 & \textbf{16.0} & 20.6 & 18.9 & 16.0 & 18.5 & 25.5 & \textbf{36.0} & 26.7 & 16.0 & 0.0 & 14.4 & 21.9 & 15.2 & 21.0 \\
TIB-L
& \textbf{60.0} & 14.4 & 20.0 & 15.3 & 29.3 & \textbf{16.0} & 27.4 & 21.7 & 28.0 & 22.5 & 16.0 & 8.0 & 15.1 & 15.1 & 0.0 & 13.6 & 19.3 & 14.4 & 18.8 \\
\hdashline
GROW-F &
59.2 & \textbf{40.8} & \textbf{36.0} & \textbf{40.0} & 36.7 & 0.0 & 31.4 & 30.9 & 40.0 & 32.0 & 22.0 & 20.0 & 21.8 & 20.0 & 0.0 & 18.0 & 30.1 & 19.2 & 28.6\Tstrut\\
GROW-L & 
59.2 & 40.0 & 32.0 & 38.7 & 29.3 & \textbf{16.0} & 27.4 & \textbf{35.4} & 40.0 & \textbf{36.0} & 25.5 & 28.0 & 25.8 & 22.2 & 0.0 & 20.0 & 30.5 & \textbf{23.2} & 29.6 \\
GROW-EWC &
59.2 & 38.4 & \textbf{36.0} & 38.0 & 25.3 & 4.0 & 22.3 & 21.1 & 40.0 & 23.5 & 26.0 & 12.0 & 24.4 & 28.9 & 0.0 & 26.0 & 28.0 & 18.4 & 26.8 \\
\hline
TIB
& \textbf{60.0} & 16.0 & 28.0 & 18.0 & 
           28.7 & 0.0 & 24.6 & 
           19.4 & \textbf{56.0} & 24.0 & 
           18.0 & 8.0 & 16.9 & 
           25.3 & 0.0 & 22.8 & 
           21.5 & 18.4 & 21.3\Tstrut\\
iManip &
\textbf{60.0} & 33.6 & 28.0 & 32.7 & 24.0 & \textbf{16.0} & 22.9 & 16.0 & 24.0 & 17.0 & 22.0 & 12.0 & 20.9 & 16.0 & 0.0 & 14.4 & 22.3 & 16.0 & 21.6 \\
GROW & 
59.2 & 40.0 & \textbf{36.0} & 39.3 & 
           \textbf{41.3} & \textbf{16.0} & \textbf{37.7} & 
           25.1 & 28.0 & 25.5 & 
           \textbf{33.5} & 16.0 & \textbf{31.6} & 
           \textbf{30.2} & 0.0 & \textbf{27.2} & 
           \textbf{34.0} & 19.2 & \textbf{32.3}\\
\hline
\end{tabular}
}
\label{a_tab:grow_update}
\vspace{-5pt}
\end{table*}

\subsubsection{GROW's Components with TIB Baseline}
We extend the key components with vanilla TIB baselines under different base and incremental iterations to form additional ablation studies as in Table~\ref{a_tab:tib+grow}. Specifically, all runs tune only the action decoder during the incremental stages, which means our addition $\mathcal{L}_\mathrm{dh}$ and action chain predictor are only functional during the base pre-training stage. Even so, adding either $\mathcal{L}_\mathrm{dh}$ or the action chain predictor can lead to a non-trivial improvement of more than $6.0\%$ and $5.4\%$ when incrementally updating with 10k and 50k iterations, respectively. This indicates the robustness of the key components in GROW under various incremental settings. 

\subsubsection{GROW with Different Update Mechanisms}
We further combine GROW with various updating mechanisms, including parameter freezing, universal LoRA, and EWC using Fisher information regularization. As shown in Table~\ref{a_tab:grow_update}, using each updating mechanism with GROW can lead to SOTA performance compared with previous TIB and iManip. One interesting observation is that the universal LoRA version of GROW (GROW-L) achieves the best average new task comprehension across all variants, surpassing vanilla GROW by $4.0\%$ on average. Nevertheless, our stage-wise LoRA with re-calibration strikes a better balance between old skill preservation and new skill comprehension, and therefore we adopt this strategy for all other experiments. Additionally, naively combining LoRA or freezing the PerceiverIO cannot achieve noticeable improvements on TIB, which demonstrates the improvement brought by the key components of GROW. 
\begin{table}[t]
\centering
\huge
\setlength{\tabcolsep}{4pt}
\caption{\textcolor{black}{Stage-wise performance ablation studies in Table~\ref{tab:main_ablation} with corresponding ablation IDs.}}
\vspace{-3pt}
\resizebox{\textwidth}{!}{
\begin{tblr}{ 
        colspec             = {c | c | c *{25}{c}},
        cell{1}{1,2} = {r=2}{l},
        cell{1}{3, 4, 8, 12, 16, 20, 24}        = {r=2}{c}, 
        cell{1}{5, 9, 13, 17, 21, 25} = {c=3}{c},
        cell{3}{1}         = {r=7}{c},
        cell{10}{1}        = {r=9}{c}
    }
\hline
& ID & Base & & Stage 1 & & & & Stage 2 & & & & Stage 3 & & & & Stage 4 & & & & Stage 5 & & & & Avg \\
\hline
& & & & Old & New & All & & Old & New & All & & Old & New & All 
   & & Old & New & All & & Old & New & All & & Old & New & All\Tstrut\\
\hline
\rotatebox[origin=c]{90}{Base + Incremental} 
& 1 & 47.2 & & 34.4 & 24.0 & 32.7 & & 
        12.0 & 32.0 & 14.9 & & 
        6.9 & 80.0 & 16.0 & & 
        17.5 & 8.0 & 16.4 & & 
        24.0 & 4.0 & 22.0 & & 
        19.0 & 29.6 & 20.4 \\
        
& 2 & 52.0 & & 26.4 & 28.0 & 26.7 & & 
        28.0 & 8.0 & 25.1 & &
        21.7 & 24.0 & 22.0 & &
        14.0 & 8.0 & 13.3 & &
        20.4 & 0.0 & 18.4 & &
        22.1 & 13.6 & 21.1\\
& 3 & 53.0 & & 32.0 & 16.0 & 29.3 & & 
        25.3 & 12.0 & 23.4 & & 
        22.9 & 20.0 & 22.5 & & 
        22.0 & 8.0 & 20.4 & & 
        20.0 & 0.0 & 18.0 & & 
        24.4 & 11.2 & 22.7\\
& 4 & 58.4 & & 28.0 & 4.0 & 24.0 & &
        30.7 & 0.0 & 26.3 & &
        35.4 & 24.0 & 34.0 & &
        29.5 & 8.0 & 27.1 & &
        16.9 & 0.0 & 15.2 & &
        28.1 & 7.2 & 25.3 \\
& 7 & 54.4 & & 31.2 & 36.0 & 32.0 & & 
        28.7 & 12.0 & 26.3 & & 
        26.3 & 40.0 & 28.0 & & 
        17.5 & 4.0 & 16.0 & & 
        18.7 & 0.0 & 16.8 & & 
        24.5 & 18.4 & 23.8\\
& 8 & 55.3 & & 29.6 & 24.0 & 28.7 & & 
        30.7 & 8.0 & 27.4 & & 
        25.1 & 36.0 & 26.5 & & 
        20.5 & 8.0 & 19.1 & & 
        19.1 & 0.0 & 17.2 & & 
        25.0 & 15.2 & 23.8 \\
& 9 & 48.8 & & 36.0 & 12.0 & 32.0 & & 
        32.7 & 4.0 & 28.6 & & 
        34.3 & 24.0 & 33.0 & & 
        23.5 & 8.0 & 21.8 & & 
        16.9 & 0.0 & 15.2 & & 
        28.7 & 9.6 & 26.1\\
\hline
\rotatebox[origin=c]{90}{Incremental}
& 1 & 59.2 & & 39.2 & 28.0 & 37.3 & & 
        33.3 & 4.0 & 29.1 & & 
        21.1 & 28.0 & 22.0 & & 
        25.0 & 8.0 & 23.1 & & 
        20.9 & 0.0 & 18.8 & & 
        27.9 & 13.6 & 26.1\\
& 2 & 59.2 & & 36.8 & 52.0 & 39.3 & & 
        38.7 & 0.0 & 33.1 & & 
        33.1 & 36.0 & 33.5 & & 
        30.5 & 16.0 & 28.9 & & 
        25.3 & 0.0 & 22.8 & & 
        32.9 & 20.8 & 31.5\\
& 3 & 59.2 & & 36.8 & 40.0 & 37.3 & & 
        33.3 & 16.0 & 30.9 & & 
        26.3 & 36.0 & 27.5 & & 
        17.5 & 24.0 & 18.2 & & 
        20.9 & 4.0 & 19.2 & & 
        27.0 & 24.0 & 26.6\\
& 4 & 59.2 & & 12.8 & 40.0 & 17.3 & &
        5.3 & 0.0 & 4.6 & &
        6.3 & 44.0 & 11.0 & &
        9.5 & 0.0 & 8.4 & &
        5.3 & 4.0 & 5.2 & &
        7.8 & 17.6 & 9.3 \\
& 5 & 59.2 & & 25.6 & 24.0 & 25.3 & &
        33.3 & 0.0 & 28.6 & &
        17.7 & 16.0 & 17.5 & &
        14.5 & 12.0 & 14.2 & &
        18.2 & 0.0 & 16.4 & &
        21.9 & 10.4 & 20.4\\

& 6 & 59.2 & & 39.2 & 20.0 & 36.0 & & 
        33.3 & 12.0 & 30.3 & & 
        34.3 & 20.0 & 32.5 & & 
        19.5 & 16.0 & 19.1 & & 
        20.4 & 0.0 & 18.4 & & 
        29.3 & 13.6 & 27.3\\
& 7 & 59.2 & & 40.8 & 40.0 & 40.7 & & 
        31.3 & 8.0 & 28.0 & & 
        33.7 & 12.0 & 31.0 & & 
        26.5 & 16.0 & 25.3 & & 
        24.4 & 0.0 & 22.0 & & 
        31.3 & 15.2 & 29.4\\
& 8 & 59.2 & & 40.0 & 16.0 & 36.0 & & 
        38.0 & 4.0 & 33.1 & & 
        3.4 & 44.0 & 8.5 & & 
        30.0 & 20.0 & 28.9 & & 
        22.7 & 0.0 & 20.4 & & 
        26.8 & 16.8 & 25.4\\
& 9 & 59.2 & & 23.2 & 28.0 & 24.0 & & 
        30.0 & 0.0 & 25.7 & & 
        25.1 & 24.0 & 25.0 & &
        23.5 & 24.0 & 23.6 & & 
        19.6 & 0.0 & 17.6 & & 
        24.3 & 15.2 & 23.2\\
\hline
\end{tblr}}\label{a_tab:stage_wise_ablation}
\end{table}

\begin{figure}[t]
    \centering
    \includegraphics[width=\linewidth]{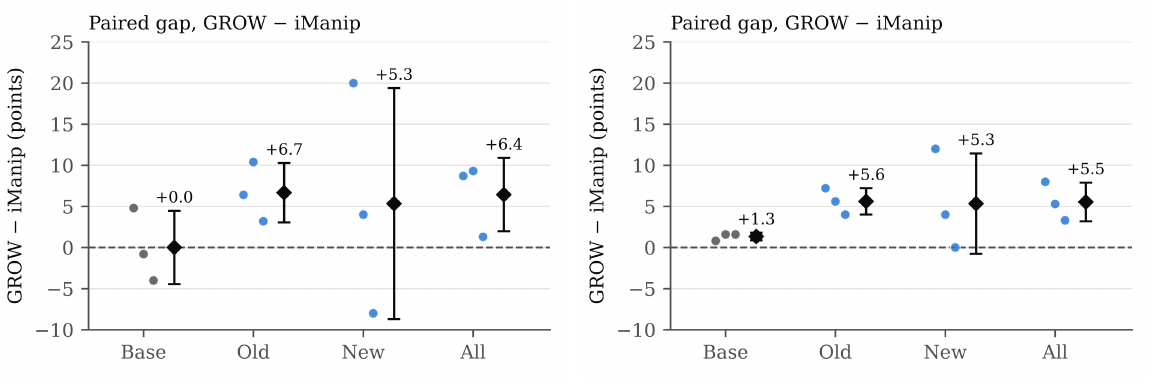}
    \caption{\textcolor{black}{Paired comparison of GROW and iManip under full randomness in B5-1N1. Runs are conducted under the default 100k+50k iterations, involving pairs of training and testing data generated using 3 different random seeds.}}
    \label{fig:a_full_random}
    \vspace{-10pt}
\end{figure}
\textcolor{black}{\subsubsection{Details of Uncertainty Estimation}\label{a_sec:uncertainty}
Table~\ref{a_tab:ic_only_uncertainty} reports the stage-wise incremental-only results summarized in Figure 5. Figure~\ref{fig:a_full_random} further visualizes the paired performance gap between GROW and iManip across three fully independent B5-1N1 runs on PerAct (left) and GNFactor (right). Despite noticeable seed-dependent variation, GROW consistently yields positive gains in old-skill retention and overall performance, with average improvements of (+6.7/+6.4) points on PerAct and (+5.6/+5.5) points on GNFactor for Old/All, respectively. In contrast, the base-performance gaps remain negligible, indicating that the improvements mainly emerge after incremental learning rather than from systematically stronger initial checkpoints. The larger variance on New reflects the sensitivity of unseen-skill acquisition to different task demonstrations and evaluation seeds.}

\subsubsection{Stage-wise Results of Ablation Studies}\label{a_sec:stage_wise_ablation}
We further supplement the stage-wise results in Table~\ref{a_tab:stage_wise_ablation} that lead to the overall ablation studies in Table~\ref{tab:main_ablation}.

\begin{table}[t]
\centering
\caption{\textcolor{black}{Stage-wise performance of incremental-only randomness analysis under B5-5N1 on PerAct, GNFactor, and ManiGaussian (mean $\pm$ sample std over three seeds). For each backbone, all runs share the same base checkpoint, training data, and evaluation scenes; seeds vary incremental training and rollout. Avg is computed per seed over the five stages, then averaged. Bold denotes the better mean in each column.}}
\label{a_tab:ic_only_uncertainty}
\small
\resizebox{.9\textwidth}{!}{\begin{tblr}{
  colspec = {l *{9}{c}},
  colsep = 3.2pt,
  rowsep = 0.6pt,
  row{5,9} = {bg=gray!15},
}
\toprule
\SetCell[c=10]{c} \textbf{PerAct} & & & & & & & & & \\
\midrule
\SetCell[r=2]{l} Method & \SetCell[c=3]{c} Stage 1 & & & \SetCell[c=3]{c} Stage 2 & & & \SetCell[c=3]{c} Stage 3 & & \\
\cmidrule[lr]{2-4} \cmidrule[lr]{5-7} \cmidrule[lr]{8-10}
 & Old & New & All & Old & New & All & Old & New & All \\
\midrule
iManip & 27.2{\scriptsize$\pm$4.2} & 12.0{\scriptsize$\pm$4.0} & 24.7{\scriptsize$\pm$3.5} & 24.5{\scriptsize$\pm$1.7} & 6.7{\scriptsize$\pm$4.6} & 21.9{\scriptsize$\pm$0.9} & 16.4{\scriptsize$\pm$3.5} & 10.7{\scriptsize$\pm$6.1} & 15.7{\scriptsize$\pm$3.3} \\
GROW & \textbf{33.1}{\scriptsize$\pm$1.2} & \textbf{25.3}{\scriptsize$\pm$9.2} & \textbf{31.8}{\scriptsize$\pm$1.0} & \textbf{28.2}{\scriptsize$\pm$1.0} & \textbf{13.3}{\scriptsize$\pm$8.3} & \textbf{26.1}{\scriptsize$\pm$1.8} & \textbf{19.2}{\scriptsize$\pm$2.3} & \textbf{13.3}{\scriptsize$\pm$6.1} & \textbf{18.5}{\scriptsize$\pm$2.8} \\
\midrule[0.08em]
\SetCell[r=2]{l} Method & \SetCell[c=3]{c} Stage 4 & & & \SetCell[c=3]{c} Stage 5 & & & \SetCell[c=3]{c} Avg & & \\
\cmidrule[lr]{2-4} \cmidrule[lr]{5-7} \cmidrule[lr]{8-10}
 & Old & New & All & Old & New & All & Old & New & All \\
\midrule
iManip & \textbf{16.8}{\scriptsize$\pm$2.8} & \textbf{14.7}{\scriptsize$\pm$4.6} & \textbf{16.6}{\scriptsize$\pm$2.0} & \textbf{17.0}{\scriptsize$\pm$6.2} & \textbf{0.0}{\scriptsize$\pm$0.0} & \textbf{15.3}{\scriptsize$\pm$5.5} & 20.4{\scriptsize$\pm$2.4} & 8.8{\scriptsize$\pm$1.6} & 18.8{\scriptsize$\pm$2.3} \\
GROW & 15.8{\scriptsize$\pm$2.0} & 13.3{\scriptsize$\pm$6.1} & 15.6{\scriptsize$\pm$1.2} & 15.6{\scriptsize$\pm$2.3} & \textbf{0.0}{\scriptsize$\pm$0.0} & 14.0{\scriptsize$\pm$2.1} & \textbf{22.4}{\scriptsize$\pm$1.4} & \textbf{13.1}{\scriptsize$\pm$2.8} & \textbf{21.2}{\scriptsize$\pm$1.2} \\
\bottomrule
\end{tblr}}

\vspace{7pt}
\resizebox{.9\textwidth}{!}{\begin{tblr}{
  colspec = {l *{9}{c}},
  colsep = 3.2pt,
  rowsep = 0.6pt,
  row{5,9} = {bg=gray!15},
}
\toprule
\SetCell[c=10]{c} \textbf{GNFactor} & & & & & & & & & \\
\midrule
\SetCell[r=2]{l} Method & \SetCell[c=3]{c} Stage 1 & & & \SetCell[c=3]{c} Stage 2 & & & \SetCell[c=3]{c} Stage 3 & & \\
\cmidrule[lr]{2-4} \cmidrule[lr]{5-7} \cmidrule[lr]{8-10}
 & Old & New & All & Old & New & All & Old & New & All \\
\midrule
iManip & 27.5{\scriptsize$\pm$5.6} & 20.0{\scriptsize$\pm$6.9} & 26.2{\scriptsize$\pm$5.7} & 25.1{\scriptsize$\pm$6.6} & \textbf{6.7}{\scriptsize$\pm$11.5} & 22.5{\scriptsize$\pm$5.2} & \textbf{16.4}{\scriptsize$\pm$2.3} & \textbf{25.3}{\scriptsize$\pm$6.1} & \textbf{17.5}{\scriptsize$\pm$1.3} \\
GROW & \textbf{42.4}{\scriptsize$\pm$3.7} & \textbf{21.3}{\scriptsize$\pm$10.1} & \textbf{38.9}{\scriptsize$\pm$2.1} & \textbf{29.8}{\scriptsize$\pm$3.3} & 4.0{\scriptsize$\pm$4.0} & \textbf{26.1}{\scriptsize$\pm$3.3} & 15.8{\scriptsize$\pm$3.7} & 10.7{\scriptsize$\pm$12.2} & 15.2{\scriptsize$\pm$3.2} \\
\midrule[0.08em]
\SetCell[r=2]{l} Method & \SetCell[c=3]{c} Stage 4 & & & \SetCell[c=3]{c} Stage 5 & & & \SetCell[c=3]{c} Avg & & \\
\cmidrule[lr]{2-4} \cmidrule[lr]{5-7} \cmidrule[lr]{8-10}
 & Old & New & All & Old & New & All & Old & New & All \\
\midrule
iManip & 23.3{\scriptsize$\pm$2.8} & 14.7{\scriptsize$\pm$2.3} & 22.4{\scriptsize$\pm$2.3} & 19.1{\scriptsize$\pm$0.8} & \textbf{0.0}{\scriptsize$\pm$0.0} & 17.2{\scriptsize$\pm$0.7} & 22.3{\scriptsize$\pm$1.8} & \textbf{13.3}{\scriptsize$\pm$1.2} & 21.2{\scriptsize$\pm$1.4} \\
GROW & \textbf{24.5}{\scriptsize$\pm$3.3} & \textbf{17.3}{\scriptsize$\pm$11.5} & \textbf{23.7}{\scriptsize$\pm$3.4} & \textbf{22.5}{\scriptsize$\pm$0.7} & \textbf{0.0}{\scriptsize$\pm$0.0} & \textbf{20.3}{\scriptsize$\pm$0.6} & \textbf{27.0}{\scriptsize$\pm$1.6} & 10.7{\scriptsize$\pm$6.2} & \textbf{24.8}{\scriptsize$\pm$1.0} \\
\bottomrule
\end{tblr}}

\vspace{7pt}
\resizebox{.9\textwidth}{!}{\begin{tblr}{
  colspec = {l *{9}{c}},
  colsep = 3.2pt,
  rowsep = 0.6pt,
  row{5,9} = {bg=gray!15},
}
\toprule
\SetCell[c=10]{c} \textbf{ManiGaussian} & & & & & & & & & \\
\midrule
\SetCell[r=2]{l} Method & \SetCell[c=3]{c} Stage 1 & & & \SetCell[c=3]{c} Stage 2 & & & \SetCell[c=3]{c} Stage 3 & & \\
\cmidrule[lr]{2-4} \cmidrule[lr]{5-7} \cmidrule[lr]{8-10}
 & Old & New & All & Old & New & All & Old & New & All \\
\midrule
iManip & 29.1{\scriptsize$\pm$4.1} & 20.0{\scriptsize$\pm$10.6} & 27.6{\scriptsize$\pm$5.0} & 23.8{\scriptsize$\pm$5.0} & 12.0{\scriptsize$\pm$6.9} & 22.1{\scriptsize$\pm$3.5} & 18.5{\scriptsize$\pm$3.4} & \textbf{29.3}{\scriptsize$\pm$4.6} & 19.8{\scriptsize$\pm$3.3} \\
GROW & \textbf{40.5}{\scriptsize$\pm$4.8} & \textbf{33.3}{\scriptsize$\pm$4.6} & \textbf{39.3}{\scriptsize$\pm$3.4} & \textbf{37.3}{\scriptsize$\pm$4.3} & \textbf{16.0}{\scriptsize$\pm$0.0} & \textbf{34.3}{\scriptsize$\pm$3.7} & \textbf{26.3}{\scriptsize$\pm$2.0} & 24.0{\scriptsize$\pm$14.4} & \textbf{26.0}{\scriptsize$\pm$0.5} \\
\midrule[0.08em]
\SetCell[r=2]{l} Method & \SetCell[c=3]{c} Stage 4 & & & \SetCell[c=3]{c} Stage 5 & & & \SetCell[c=3]{c} Avg & & \\
\cmidrule[lr]{2-4} \cmidrule[lr]{5-7} \cmidrule[lr]{8-10}
 & Old & New & All & Old & New & All & Old & New & All \\
\midrule
iManip & 18.0{\scriptsize$\pm$3.8} & 8.0{\scriptsize$\pm$4.0} & 16.9{\scriptsize$\pm$3.8} & 16.3{\scriptsize$\pm$1.8} & 1.3{\scriptsize$\pm$2.3} & 14.8{\scriptsize$\pm$1.8} & 21.1{\scriptsize$\pm$1.7} & 14.1{\scriptsize$\pm$2.0} & 20.2{\scriptsize$\pm$1.8} \\
GROW & \textbf{29.8}{\scriptsize$\pm$3.2} & \textbf{20.0}{\scriptsize$\pm$6.9} & \textbf{28.7}{\scriptsize$\pm$2.7} & \textbf{25.8}{\scriptsize$\pm$5.0} & \textbf{2.7}{\scriptsize$\pm$4.6} & \textbf{23.5}{\scriptsize$\pm$4.5} & \textbf{31.9}{\scriptsize$\pm$1.8} & \textbf{19.2}{\scriptsize$\pm$5.6} & \textbf{30.4}{\scriptsize$\pm$1.8} \\
\bottomrule
\end{tblr}}
\end{table}

\begin{figure}[t]
    \centering
    \includegraphics[width=\linewidth]{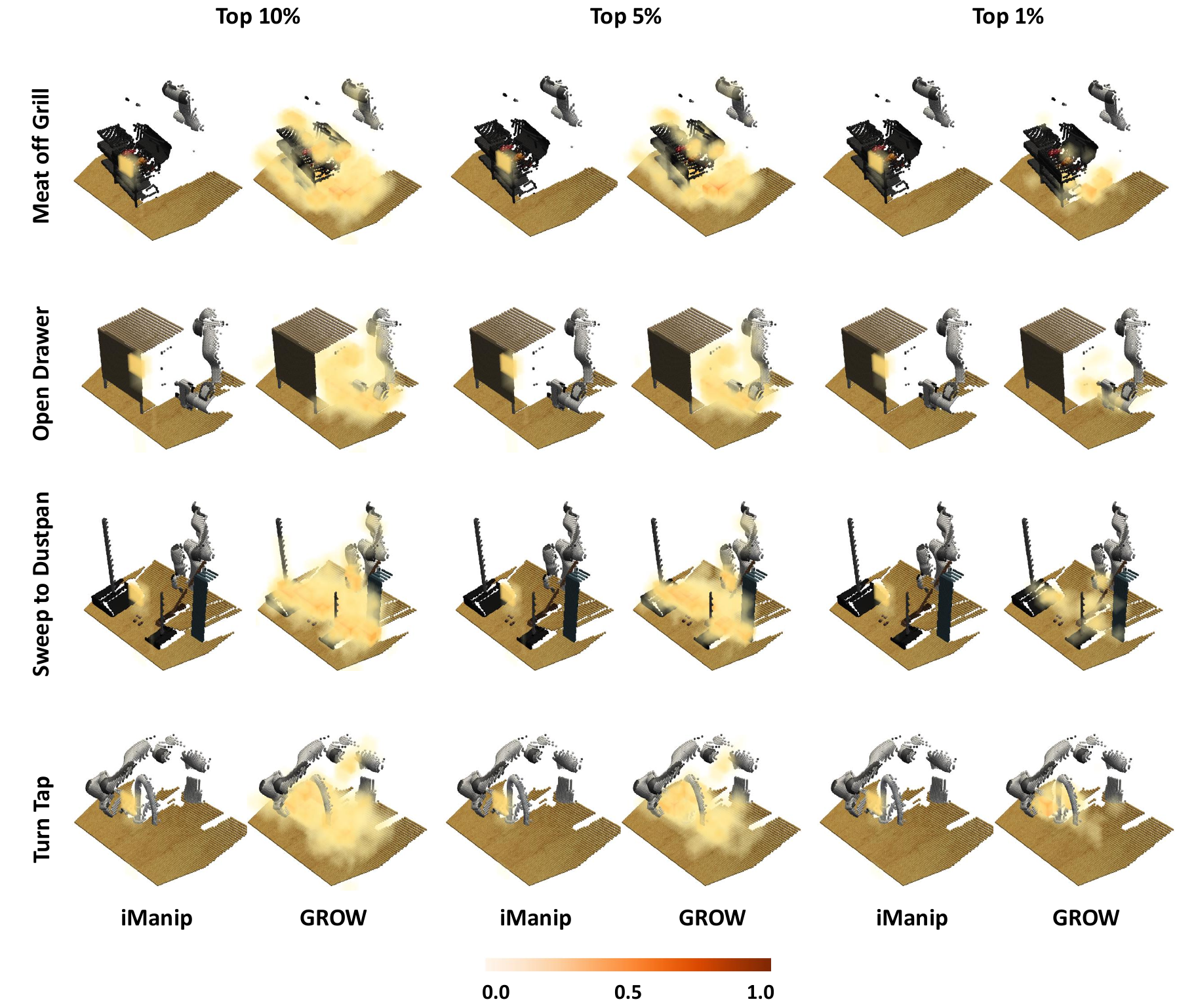}
    \caption{More comparison of attention heatmaps between iManip and GROW. All results are from the checkpoints of Stage 5 .}
    \label{fig:a_attn_heatmap}
    \vspace{-10pt}
\end{figure}

\subsection{Qualitative Visualization}
\subsubsection{More Comparison of Attention Heatmaps}
Figure~\ref{fig:a_attn_heatmap} supplements more heatmap visualizations comparing GROW with iManip. We specifically visualize the heatmap of the final-stage checkpoints against their attention focus with regard to the base tasks (task IDs 2-5 as in Table~\ref{a_tab:task_variations}). For all shown tasks and threshold levels, GROW exhibits more compact and semantically meaningful activation patterns, remaining concentrated around the objects and regions directly involved in the skill execution. By contrast, iManip often allocates a noticeable portion of its high-response attention to the upper corner of the voxel space, away from the actual manipulation region. This phenomenon indicates that, after repeated incremental updates, GROW better preserves task-relevant perception grounding, while iManip suffers more from attention drift.

\subsection{Simulation Rollout Videos}
To further qualitatively evaluate long-term skill retention, we provide 10 simulation rollout videos collected using the final Stage~5 checkpoints of iManip and GROW in our supplementary materials. All videos are evaluated on the base skills that were learned before incremental adaptation, allowing direct comparison of how well previously acquired behaviors are preserved after the complete five-stage learning sequence. The paired rollouts highlight differences in execution stability, task-relevant interaction, and failure modes between the two methods. In particular, GROW more consistently maintains coherent skill execution on these previously learned tasks, while iManip exhibits more frequent degradation after long-sequence incremental updates. These qualitative results complement the quantitative evaluation by illustrating how differences in skill-level retention manifest during actual policy rollout.

\section{Hyper-Parameter Analysis}
\begin{table}[t]
\centering
\caption{Hyper-parameter analysis of GROW.}
\label{a_tab:hyperparam}
\setlength{\tabcolsep}{5pt}
\resizebox{.5\textwidth}{!}{
\begin{tblr}{
    colspec = {c c c c c},
    row{1} = {font=\bfseries},
    cell{2,6,10,14}{1} = {r=4}{c},
    hline{1,Z} = {1pt},
    hline{2} = {0.6pt},
    hline{6,10,14} = {0.6pt},
    rows = {valign=m},
    column{3-5} = {halign=c},
}
\textbf{Hyper-parameter} &
\textbf{Setting} &
\textbf{Old} &
\textbf{New} &
\textbf{All} \\

$\lambda_{\mathrm{next}}$
& $0.03$ & \textbf{45.6} & \textbf{36.0} & \textbf{44.0} \\
& $0.05$ & 40.0 & 20.0 & 36.7 \\
& $0.07$ & 33.6 & 32.0 & 33.3 \\
& $0.10$ & 32.8 & 28.0 & 32.0 \\

$\lambda_{\mathrm{goal}}$
& $0.03$ & 30.4 & 32.0 & 30.7 \\
& $0.05$ & \textbf{45.6} & \textbf{36.0} & \textbf{44.0} \\
& $0.07$ & 36.0 & 36.0 & 36.0 \\
& $0.10$ & 37.6 & 4.0 & 32.0 \\

$\lambda_{\mathrm{chain}}$
& $0.003$ & 35.2 & 32.0 & 34.7 \\
& $0.005$ & \textbf{37.6} & 28.0 & \textbf{36.0} \\
& $0.007$ & 34.4 & \textbf{36.0} & 34.7 \\
& $0.010$ & 33.6 & 24.0 & 32.0 \\

$\sigma$
& $1$ & 35.2 & 12.0 & 31.3 \\
& $2$ & 28.8 & 28.0 & 28.7 \\
& $4$ & \textbf{34.4} & \textbf{36.0} & \textbf{34.7} \\
& $5$ & 33.6 & 24.0 & 32.0 \\

\hline
GROW & - & 40.0 & 36.0 & 39.3

\end{tblr}}
\end{table}

Table~\ref{a_tab:hyperparam} studies the sensitivity of GROW to the loss coefficients of next-step grounding \(\lambda_{\mathrm{next}}\), goal grounding \(\lambda_{\mathrm{goal}}\), and action-chain prediction \(\lambda_{\mathrm{chain}}\), together with the Gaussian variance \(\sigma\) used for spatial grounding. \textcolor{black}{It is a later, post-hoc sensitivity analysis on the main setting. The default loss coefficients were fixed early in the project by checking a small number of settings on a shortened schedule (50k base + 10k incremental iterations, first incremental stage only; the setting of Table~\ref{a_tab:tib+grow}), evaluated on the same evaluation episodes; no dedicated validation split was used.}

Overall, GROW remains reasonably robust across a broad range of settings, although excessively large grounding or chain weights can noticeably degrade the balance between old-skill retention and new-skill acquisition. Notably, \(\lambda_{\mathrm{next}}=0.03\) and \(\lambda_{\mathrm{goal}}=0.05\) achieve higher performance than our default configuration in this stage-wise analysis, reaching \(44.0\%\) overall success compared with \(39.3\%\). \textcolor{black}{Since the defaults were fixed before this analysis, we did not adopt these configurations and retained the default hyperparameters throughout the main experiments.} This analysis is conducted using the ManiGaussian backbone, while the same default hyperparameters are directly adopted for PerAct and GNFactor without backbone-specific tuning. Their reported performance may therefore be further improved with dedicated hyperparameter optimization.

\section{Real-World Setup}
In this section, we introduce the setup of our real-world experiments, including hardware setup, task definition and postprocessing, evaluation rollouts, and the qualitative results. 

\begin{figure}[t]
    \centering
    \includegraphics[width=0.7\linewidth]{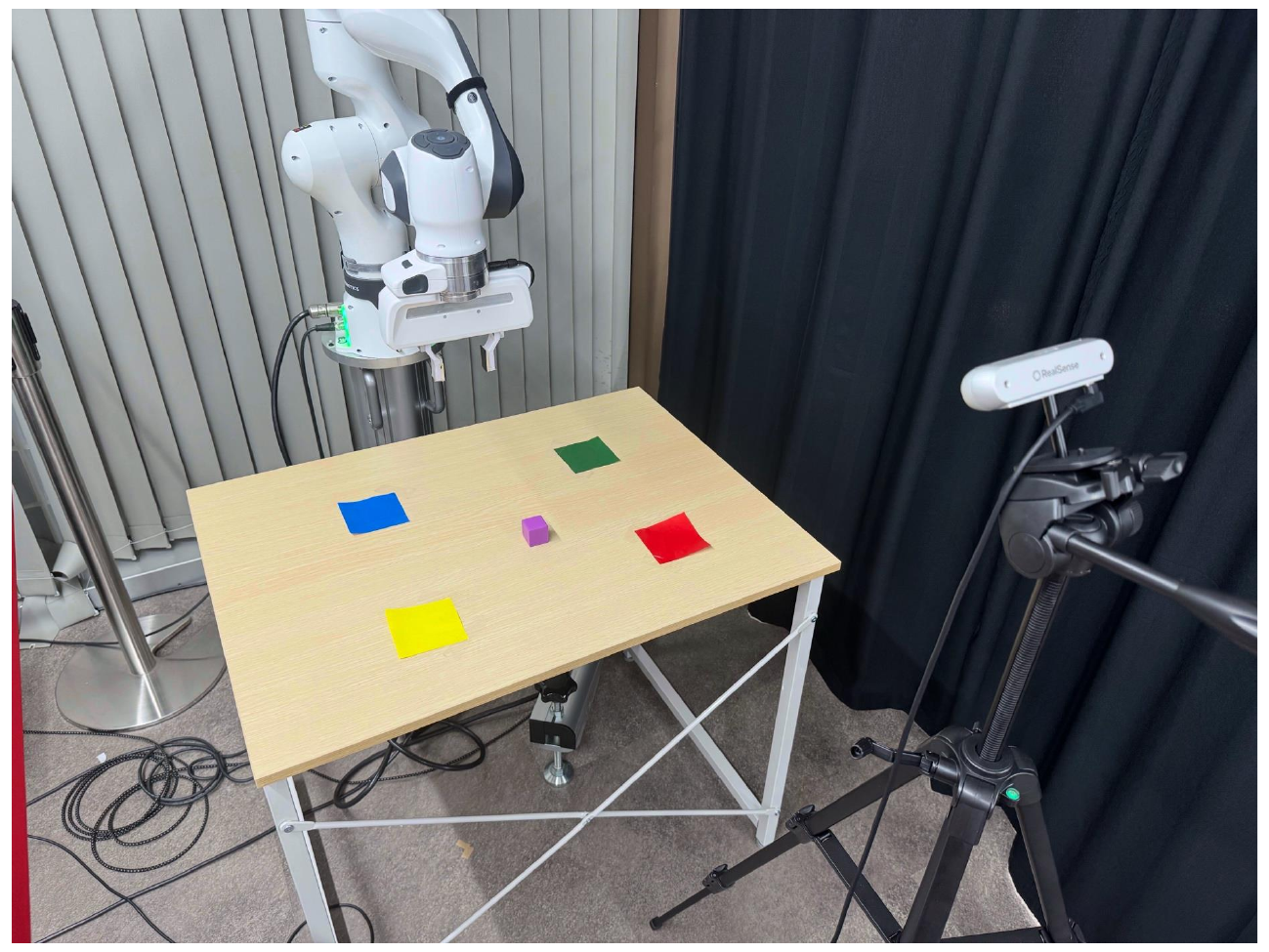}
    \caption{Hardware setup for real-world experiments.}
    \label{a_fig:hardware}
\end{figure}

\subsection{Hardware setup}
The real-world platform consists of a seven-degree-of-freedom Franka FR3 manipulator equipped with a parallel-jaw gripper. \textcolor{black}{Specifically, we adopt the Franka Research 3 (FR3) arm (Arm v1), produced in September, 2024. The model number of the arm controller is 10070380, manufactured in August 2024. The gripper fingertip is 3D printed, which shares a similar shape with the official Franka Hand’s, but extends the length of the contact surface from 1.5cm to 3cm. Anti-slip mats are pasted on both fingertips to increase the reliability of the gripper.} Manipulation takes place in a tabletop workspace containing the objects required for block stacking, placing items into a basket, and sliding a block onto a colored target. The robot base provides the common reference frame for describing the workspace, camera geometry, and end-effector states.

\textcolor{black}{Visual observations are obtained from a fixed, external Intel RealSense D455 RGB-D camera (TA No. K83122-212, a recent product whose batch and manufacture date have not yet been documented) overlooking the manipulation area.} The experiments use a single observation viewpoint only. The D455 camera is configured to provide RGB images at \(640\times480\) pixels and depth images at \(848\times480\) pixels, both at a nominal frequency of 15 Hz. Camera synchronization, depth-to-color alignment, and colored point-cloud generation are enabled. We perform eye-on-base calibration between the external camera
and the robot using the \texttt{easy\_handeye} package%
\footnote{\url{https://github.com/IFL-CAMP/easy_handeye}}. 

Data collection is done by teleoperation with Meta Quest 2, with most of the code adopted from \texttt{franka\_vr}\footnote{\url{https://github.com/ZorAttC/franka_vr}}. To facilitate more effective and convenient teleoperation, the controller buttons provide translation-only, rotation-only, yaw-only, and roll/pitch-only modes. Switching these constraints re-establishes the reference at the current robot and controller poses to avoid returning to an older reference. A clutch command similarly re-establishes the reference without commanding arm movement. A precision mode reduces both translational and angular velocity limits to $35\%$ of their normal values. Together, we can operate the FR3 to do some precise movement, manipulating small objects such as blocks and a toy pear. We attach a teleoperation video in the supplementary materials for reference. Robot and camera streams are recorded separately as ROS bags. Robot recordings include arm joint states, gripper joint states, and dynamic and static transforms; camera recordings contain the enabled sensor streams. Recorded timestamps support subsequent temporal alignment and reconstruction of end-effector trajectories. This preserves measured robot states independently of the operator’s requested targets. \textcolor{black}{The software stack consists of libfranka 0.15.0 and franka\_ros2 v2.0.4 on FR3 system version 5.7.2 (robot server 9, gripper server 3), ROS~2 Humble (ros-base 0.10.0, rclcpp 16.0.19) on Ubuntu 22.04, and librealsense 2.58.4 with realsense2\_camera 4.58.4 for the D455 (firmware 5.17.3.10).}

\subsection{Task Definition and Postprocessing}
\textbf{Task Definition and Variations.}
Our real-world experiments involve three tasks, including \texttt{put\_items\_in\_baskets}, \texttt{stack\_blocks}, and \texttt{slide\_blocks\_to\_regions}. Each collected demonstration has a corresponding record in our $\texttt{*.csv}$ table in the supplementary materials, including the linguistic instruction, environmental descriptions, and ROS bag ID. Specifically, for \texttt{stack\_blocks} the 20 demonstrations cover different ordered color pairs among red, blue, and green blocks, requiring the policy to distinguish both the source and target block while operating in a consistent two-block scene. For \texttt{put\_items\_in\_baskets}, the demonstrations vary both object identity (red/green apple, strawberry, and pear) and target basket color, while progressively increasing scene complexity from one object with two baskets, to two objects with two baskets, and finally four objects with three baskets. For \texttt{slide\_blocks\_to\_regions}, the demonstrations vary the instructed target color as well as the set of candidate colored regions in the scene, including combinations of red, green, blue, yellow, orange, and black targets.

\textbf{Postprocessing and Training.}
The postprocessing procedures extract the raw ROS bags into keyframes used to train PerAct models for each incremental method. All geometric observations and action targets are first expressed in the robot base frame. Camera-frame points are transformed into the base frame using the calibrated camera-to-base transformation. Non-finite points and points outside the configured workspace are removed. The workspace bounds are \(x\in[0.1,1.2]\), \(y\in[-0.5,0.6]\), and \(z\in[-0.12,0.88]\) metres. Each retained observation is represented by 16,384 paired XYZ–RGB samples. Deterministic random sampling is performed without replacement when sufficient valid points are available and with replacement otherwise. The sampled coordinates and colors are stored as \(128\times128\times3\) arrays for compatibility with the downstream data interface. This arrangement is a storage convention and does not preserve the original camera-image pixel layout. 

For keyframe extraction, considering the fact that teleoperation usually involves small but repetitive local motion when targeting small objects, we adopt greedy pruning to remove consecutive and similar keyframes. Specifically, keyframes are detected from robot stops, gripper-state changes, and the terminal observation. \textcolor{black}{Stop detection uses a joint-velocity tolerance of 0.02 rad/s across the interpolated FR3 joint velocities. Redundant stopped poses are subsequently pruned using translation and rotation thresholds of 1 cm and 5°, respectively. }Gripper-change and terminal anchors are preserved during this pose-based pruning.

As for the training details, we adopt the identical hyperparameter settings and configurations as in the simulations, forming a fair re-implementation from simulation to real-world rollouts. 

\subsection{Evaluation Rollouts}
The evaluation rollouts are conducted following some universal protocols. We try our best to maintain the same environmental conditions for evaluation across all methods (e.g., object positions and lighting conditions), although there might be some inevitable minor differences across each rollout. The environment conditions are uniformly sampled from the training data with different object positions. The online prediction is generated using a single RTX 4090 GPU under a keyframe prediction paradigm: the agent captures the observation when the end-effector point reaches the previous target pose, and then predicts the next end-effector pose for execution. A manual buffer time of 3s is set between executing the predicted end-effector pose as a safety measure, during which an operator may terminate unsafe executions; predictions are never corrected, and terminated rollouts count as failures without retry. Rollouts that exceed 25 steps without completing the skill are considered failures. 

\subsection{Qualitative Results}
We include three pairs of rollout videos in the supplementary materials, each comparing iManip and GROW on the same real-world skill. These pairs were selected to qualitatively illustrate differences in execution and are not representative of the average margins in Table~\ref{tab:realworld_performance}. For \texttt{stack\_blocks}, GROW stacks the green block on the blue block, whereas iManip approaches a location slightly offset from the blue block and fails. For \texttt{put\_items\_in\_baskets}, iManip fails to localize and grasp the instructed apple, whereas GROW grasps it and places it into the corresponding basket. For \texttt{slide\_blocks\_to\_regions}, iManip remains largely inactive after initialization, whereas GROW slides the block toward the designated target. To provide a balanced view, we further include three pairs in which GROW fails. These cover similar inactive cases as iManip in \texttt{slide\_blocks\_to\_regions}, an object-order confusion in \texttt{stack\_blocks}, and a misplacement in \texttt{put\_items\_in\_baskets}, where GROW places the pear into a basket of the wrong color. These failures show that instruction grounding remains imperfect with the PerAct backbone, consistent with the limitations discussed in Sec.~\ref{sec:limitations}.

\end{document}